\documentclass[conf]{new-aiaa}
\usepackage[table]{xcolor}
\usepackage{textcomp}
\usepackage{amsmath,amsfonts}
\usepackage{graphicx}
\usepackage{tikz}
\usepackage{url}

\usepackage{colortbl}
\usepackage{booktabs}
\usepackage{array}
\usepackage{tabularx}

\newcolumntype{L}[1]{>{\raggedright\let\newline\\\arraybackslash\hspace{0pt}}m{#1}}

\makeatletter
\DeclareRobustCommand\onedot{\futurelet\@let@token\@onedot}
\def\@onedot{\ifx\@let@token.\else.\null\fi\xspace}

\usepackage[T1]{fontenc}  
\usepackage[b]{esvect}    

\makeatletter
\newlength\xvec@height%
\newlength\xvec@depth%
\newlength\xvec@width%
\newcommand{\xvec}[2][]{%
  \ifmmode%
    \settoheight{\xvec@height}{$#2$}%
    \settodepth{\xvec@depth}{$#2$}%
    \settowidth{\xvec@width}{$#2$}%
  \else%
    \settoheight{\xvec@height}{#2}%
    \settodepth{\xvec@depth}{#2}%
    \settowidth{\xvec@width}{#2}%
  \fi%
  \def\xvec@arg{#1}%
  \def\xvec@dd{:}%
  \def\xvec@d{.}%
  \raisebox{.2ex}{\raisebox{\xvec@height}{\rlap{%
    \kern.05em
    \begin{tikzpicture}[scale=1]
    \pgfsetroundcap
    \draw (.05em,0)--(\xvec@width-.05em,0);
    \draw (\xvec@width-.05em,0)--(\xvec@width-.15em, .075em);
    \draw (\xvec@width-.05em,0)--(\xvec@width-.15em,-.075em);
    \ifx\xvec@arg\xvec@d%
      \fill(\xvec@width*.45,.5ex) circle (.5pt);%
    \else\ifx\xvec@arg\xvec@dd%
      \fill(\xvec@width*.30,.5ex) circle (.5pt);%
      \fill(\xvec@width*.65,.5ex) circle (.5pt);%
    \fi\fi%
    \end{tikzpicture}%
  }}}%
  #2%
}
\makeatother

\makeatletter
\renewcommand*\env@matrix[1][\arraystretch]{%
  \edef\arraystretch{#1}%
  \hskip -\arraycolsep
  \let\@ifnextchar\new@ifnextchar
  \array{*\c@MaxMatrixCols c}}
\makeatother

\definecolor{commentcolor}{gray}{0.5}
\usepackage{algorithm}
\usepackage{algpseudocode}

\algnewcommand{\LineComment}[1]{\State \textcolor{commentcolor}{\(\triangleright\) #1}}
\algnewcommand{\To}{\textbf{to}}
\algnewcommand{\Break}{\textbf{break}}
\algnewcommand{\Continue}{\textbf{continue}}
\algnewcommand{\IIf}[1]{\State\algorithmicif\ #1\ \algorithmicthen}
\algnewcommand{\EndIIf}{\unskip}
\algnewcommand{\var}[1]{\textit{#1}}
\algnewcommand{\func}[1]{\textsc{#1}}

\usepackage[utf8]{inputenc}
\usepackage{multirow}
\usepackage{graphicx}
\usepackage{amsmath}
\usepackage[version=4]{mhchem}
\usepackage{siunitx}
\usepackage{longtable,tabularx}
\usepackage{epstopdf}
\usepackage{setspace}
\usepackage{mathtools}
\usepackage{soul}
\usepackage{grffile}
\usepackage{threeparttable}
\usepackage{gensymb}
\usepackage{todonotes}
\usepackage{diagbox}
\usepackage{graphicx}
\usepackage{subcaption} 
\usepackage{booktabs}
\usepackage{graphicx} 
\usepackage{rotating}
\usepackage{array}
\usepackage{threeparttable}
\usepackage[absolute,overlay]{textpos}
\usepackage{placeins}
\usepackage{tikz}
\usetikzlibrary{arrows.meta, positioning}

 \title{Physics-infused Learning for Aerial Manipulator in Winds and Near-Wall Environments}

\author{Yiming Zhang\footnote{Independent Researcher, zzzraymond0@gmail.com}}
\author{Junyi Geng\footnote{Assistant Professor of Aerospace Engineering, Member AIAA, jgeng@psu.edu}}
\affil{The Pennsylvania State University, University Park, PA 16802}

\begin{document}
\begin{textblock*}{15cm}(4.75cm,1cm) 
   {\Large\textcolor{red}{AIAA SciTech Forum, January 12–16 2026}\\
   \textcolor{purple}{\url{https://arc.aiaa.org/doi/10.2514/6.2026-1760}}}
\end{textblock*}

\maketitle

\begin{abstract}

Aerial manipulation (AM) expands UAV capabilities beyond passive observation to contact-based operations at high altitudes and in otherwise inaccessible environments. Although recent advances show promise, most AM systems are developed in controlled settings that overlook key aerodynamic effects.  
Simplified thrust models are often insufficient to capture the nonlinear wind disturbances and proximity-induced flow variations present in real-world environments near infrastructure, while high-fidelity CFD methods remain impractical for real-time use. Learning-based models are computationally efficient at inference, but often struggle to generalize to unseen condition. 
This paper combines both approaches by integrating a physics-based blade-element model with a learning-based residual force estimator, along with a rotor-speed allocation strategy for disturbance compensation, resulting in a unified control framework. 
The blade-element model computes per-rotor aerodynamic forces under wind and provides a refined feedforward disturbance estimate. A learning-based estimator then predicts the residual forces not captured by the model, enabling compensation for unmodeled aerodynamic effects. 
An online adaptation mechanism further updates the residual-force prediction and rotor-speed allocation jointly to reduce the mismatch between desired and realized thrust.
We evaluate this framework in both free-flight and wall-contact tracking tasks in a simulated near-wall wind environment. Results demonstrate improved disturbance estimation and trajectory-tracking accuracy over conventional approaches, enabling robust wall-contact execution under challenging aerodynamic conditions.
\end{abstract}

\section*{Nomenclature} \label{sec:nomenclature}
\begin{longtable}{ll}
\label{tab:symbols}
\( \mathbf{M} \in \mathbb{R}^{6 \times 6} \) & Generalized inertia matrix in the body frame \\
\( \mathbf{C} \in \mathbb{R}^{6 \times 6} \) & Coriolis and centrifugal matrix in the rigid-body model \\
\( \mathbf{V} \in \mathbb{R}^6 \) & Body-frame twist: linear and angular velocity \( \begin{bmatrix} \mathbf{v}^\top & \boldsymbol{\omega}^\top \end{bmatrix}^\top \) \\
\( \boldsymbol{\omega} \in \mathbb{R}^3 \) & Angular velocity of the vehicle in the body frame \\
\( \mathbf{v} \in \mathbb{R}^3 \) & Linear velocity of the vehicle in the body frame \\
\( \mathbf{R} \in \mathrm{SO}(3) \) & Rotation matrix from inertial frame to body frame \\
\( m \in \mathbb{R} \) & Mass of the vehicle \\
\( \mathbf{J} \in \mathbb{R}^{3 \times 3} \) & Inertia matrix of the vehicle in the body frame \\
\( \mathbf{g} \in \mathbb{R}^3 \) & Gravitational acceleration vector in inertial frame \\
\( \mathbf{G} \in \mathbb{R}^6 \) & Gravity wrench in body frame \\
\( \boldsymbol{\tau}_{\text{ext}} \in \mathbb{R}^6 \) & External wrench acting on the vehicle \\
\( \boldsymbol{\tau}_a \in \mathbb{R}^6 \) & Aerodynamic wrench (force and moment) \\
\( \boldsymbol{\tau}_c \in \mathbb{R}^6 \) & Contact wrench (assumed zero in free flight) \\
\( \mathbf{F}_a \in \mathbb{R}^3 \) & Total aerodynamic force from all rotors \\
\( \mathbf{T}_a \in \mathbb{R}^3 \) & Total moment from all rotors \\
\( \mathbf{F}_i \in \mathbb{R}^3 \) & Aerodynamic force from rotor \( i \) \\
\( \mathbf{T}_i \in \mathbb{R}^3 \) & Torque vector from rotor \( i \) \\
\( \mathbf{r}_i \in \mathbb{R}^3 \) & Vector from vehicle center of mass to rotor \( i \), in body frame \\
\( d_h \in \mathbb{R} \) & In-plane (horizontal) radial distance between the rotor and the vehicle’s center of mass \\
\( d_v \in \mathbb{R} \) & Out-of-plane (vertical) offset between the rotor and the vehicle’s center of mass \\
\( n_i \in \mathbb{R} \) & Propeller rotational speed of rotor \( i \) \\
\( \mathbf{v}_{\text{wind},i} \in \mathbb{R}^3 \) & Freestream wind velocity at rotor \( i \)'s hub \\
\( \mathbf{v}_{\text{prop},i} \in \mathbb{R}^3 \) & Translational velocity of rotor \( i \)'s hub \\
\( c_{\ell,1} \in \mathbb{R} \) & Linear lift--curve slope of the propeller blade airfoil section \\
\( c_{\ell,2} \in \mathbb{R} \) & Post--stall lift coefficient factor of the propeller blade airfoil section \\
\( c_d \in \mathbb{R} \) & Profile--drag scale coefficient of the propeller blade airfoil section \\
\( \alpha_0 \in \mathbb{R} \) & Stall--transition angle of the propeller blade airfoil section \\
\( \theta(\cdot) \) & Blade pitch distribution along the span \\
\( c(\cdot) \) & Blade chord distribution along the span \\
\( n_{i,\text{cmd}} \in \mathbb{R} \) & Commanded rotor speed generated by the controller \\
\( T_m \in \mathbb{R}_{>0} \) & Motor time constant capturing the first-order rotor speed delay \\
\( \boldsymbol{u} \in \mathbb{R}^6 \) & Control wrench generated by the tracking controller \\
\( \boldsymbol{\tau}_d \in \mathbb{R}^6 \) & Net disturbance wrench \\
\( \hat{\boldsymbol{\tau}}_d \in \mathbb{R}^6 \) & Estimated disturbance wrench used for compensation \\
\( \mathbf{V}_r \in \mathbb{R}^6 \) & Reference twist (desired body-frame linear and angular velocity) \\
\( \mathbf{s} \in \mathbb{R}^6 \) & Tracking error variable used in the geometric controller \\
\( K \in \mathbb{R}^{6 \times 6} \) & Positive-definite tracking gain matrix in the control law \\

\( \bar{\boldsymbol{\tau}}_a \in \mathbb{R}^6 \) & Nominal aerodynamic wrench predicted by the BEMT-based model \\
\( \bar{\mathbf{F}}_a \in \mathbb{R}^3 \) & Nominal net aerodynamic force predicted by the BEMT-based model \\
\( \bar{\mathbf{T}}_a \in \mathbb{R}^3 \) & Nominal net aerodynamic moment predicted by the BEMT-based model \\
\( \bar{\mathbf{F}}_{a,\text{lat}} \in \mathbb{R}^3 \) & Lateral (non-axial) component of the nominal aerodynamic force \\
\( \bar{F}_x,\, \bar{F}_y,\, \bar{F}_z \in \mathbb{R} \) & Components of the nominal aerodynamic force in the body frame \\
\( \hat{\mathbf{F}}_d \in \mathbb{R}^3 \) & Estimated disturbance force used in the controller \\
\( \mathbf{F} \in \mathbb{R}^3 \) & Net aerodynamic force acting on the vehicle (from dynamics model) \\
\( \mathbf{F}_{\text{sensed}} \in \mathbb{R}^3 \) & Net aerodynamic force reconstructed from sensor measurements \\
\( \boldsymbol{\phi}(\mathbf{x}_\phi) \in \mathbb{R}^3 \) & Learned residual aerodynamic force as a function of input \(\mathbf{x}_\phi\) \\
\( \mathbf{x}_\phi \in \mathbb{R}^{n_\phi} \) & Input vector to the residual network (state- and flow-related quantities) \\
\( \hat{\boldsymbol{\epsilon}} \in \mathbb{R}^3 \) & Online adaptive correction term for residual disturbances \\

\( x_a \in \mathbb{R}^4 \) & Aerodynamic parameter vector \( (c_{\ell,1}, c_{\ell,2}, c_d, \alpha_0) \) \\
\( x_a^{*} \in \mathbb{R}^4 \) & Optimal aerodynamic parameter vector learned from data \\
\( \mathbf{v}_{\text{sensed}} \in \mathbb{R}^3 \) & Measured linear velocity in the body frame \\
\( \boldsymbol{\omega}_{\text{sensed}} \in \mathbb{R}^3 \) & Measured angular velocity in the body frame \\
\( F_z \in \mathbb{R} \) & Total thrust command along the body \(z\)-axis \\
\( M_x,\, M_y,\, M_z \in \mathbb{R} \) & Commanded body-frame roll, pitch, and yaw moments \\
\( f_i \in \mathbb{R} \) & Allocated thrust for rotor \( i \) \\
\( c_{\tau f} \in \mathbb{R} \) & Torque coefficient relating rotor thrust to reactive (yaw) torque \\

\( \mathcal{L} \in \mathbb{R} \) & Training loss for the residual neural network \\
\( N \in \mathbb{N} \) & Number of training samples \\
\( B \in \mathbb{N} \) & Batch size used during neural network training \\

\( \lambda \in \mathbb{R}_{>0} \) & Regularization coefficient in the adaptation law \\
\( \mathbf{P} \in \mathbb{R}^{3 \times 3} \) & Covariance matrix for the adaptive residual disturbance estimator \\
\( \mathbf{Q} \in \mathbb{R}^{3 \times 3} \) & Process-noise gain matrix in the covariance update \\
\( \mathbf{R} \in \mathbb{R}^{3 \times 3} \) & Measurement-noise gain matrix in the covariance update \\
\( \boldsymbol{y} \in \mathbb{R}^3 \) & Measured residual force used in the adaptation law \\

\end{longtable}

\section{Introduction} \label{sec:intro}
Unmanned Aerial Vehicles (UAVs) have become ubiquitous in aerial sensing and surveillance applications due to their agility, accessibility, and ease of deployment. Despite this widespread use, most UAV applications remain limited to passive tasks such as aerial photography, mapping, or remote monitoring. However, aerial manipulation (AM), the use of UAVs equipped with a manipulation tool to physically interact with the environment, has become more feasible and practical. AM provides a new solution to contact-based tasks at high altitudes and in otherwise inaccessible or hazardous environments, such as bridge painting~\cite{he2023image}, firefighting~\cite{aydin2019use}, seed planting~\cite{ahirwar2019application} and power line de-icing~\cite{li2024investigation}.

Research in AM has seen increasing interest for tasks such as pick-and-place~\cite{appius2022raptor}, contact inspection~\cite{guo2023aerial}, and aerial writing~\cite{guo2024flying}. Efforts in this domain have spanned controller design~\cite{lee2010geometric}, trajectory planning~\cite{mellinger2011minimum}, visual servoing~\cite{he2023image}, and human-guided learning~\cite{goecks2019efficiently, he2025flying}. Despite these advancements, most existing approaches are developed and validated in controlled indoor environments. Such settings fail to capture critical real-world effects, particularly aerodynamic disturbances from strong winds, proximity-induced flow interference, and boundary layer phenomena that arise when UAVs operate near infrastructure. These limitations hinder the deployment of aerial manipulators in practical scenarios, where robustness to environmental uncertainties is essential for reliable operation.

For practical AM at high altitudes, such as bridge maintenance, power line inspection, or infrastructure cleaning, aerodynamic effects become a critical factor that cannot be ignored. Two major sources of aerodynamic disturbance are ambient wind and proximity-induced flow effects. Strong wind disturbances can impose large, uncertain forces on the vehicle~\cite{o2022neural}. Meanwhile, UAVs operating near structures are subject to proximity-induced effects, such as ground, ceiling and wall effect~\cite{tanabe2018multiple}, which can significantly change local flow conditions. These proximity effects distort the surrounding airflow, introducing nonlinear interactions that are challenging to model accurately~\cite{shi2019neural}. Moreover, wind itself becomes more complex in the vicinity of infrastructure; as it interacts with buildings or bridges, the flow field can vary sharply across space, resulting in location-dependent disturbances that further complicate prediction and control~\cite{raza2017experimental}. These combined effects, including non-uniform wind fields and proximity-driven flow redistribution, make the aerodynamic behavior of UAVs in such environments highly dynamic and difficult to model.

Traditional methods based on simplified aerodynamic models often struggle to capture these nonlinear interactions. One widely used model assumes that rotor thrust and yaw moment are quadratic functions of propeller speed~\cite{mellinger2011minimum}. Although this assumption holds under hovering conditions in still air, the relationship between propeller speed, thrust, and yaw moment can change significantly during forward flight or in the presence of wind~\cite{gill2017propeller}. For example, when a quadrotor performs aggressive maneuvers or experiences strong wind in the vicinity of infrastructure, rotor drag acting parallel to the rotor disk plane becomes non-negligible.
While computational fluid dynamics (CFD) can provide high-fidelity aerodynamic modeling~\cite{thai2019cfd}, it is often too computationally expensive for real-time applications. As an alternative, learning-based approaches can be trained offline on flight data to approximate unmodeled aerodynamic effects. Once trained, these models can predict aerodynamic forces and are lightweight enough to run in real time on embedded controllers. However, they are typically trained on free-flight data~\cite{shi2019neural} and often fail to capture the coupled dynamics that arise during physical interactions near surfaces. 

The blade element model provides another modeling option, capturing rotor-induced disturbances using blade element theory~\cite{gill2017propeller}. It relates rotor forces to the combination of free-stream wind and induced flow but does not account for other aerodynamic effects, such as body drag acting on the UAV structure. We observe that the blade element model effectively improves feedforward actuation accuracy by capturing rotor-induced aerodynamic effects, while learning-based methods excel at predicting complex, unmodeled dynamics. Motivated by the complementary strengths of these two approaches, this paper proposes a hybrid modeling and control framework that integrates a physics-based blade element model with a learning-based disturbance estimator. The blade element model provides a structured estimation of aerodynamic forces for improved feedforward control, and the learning module compensates for residual disturbances that are difficult to capture analytically, particularly under complex environmental interactions.
Based on the learned model, we then develop an aerodynamic disturbance-rejection control approach that enhances rotorcraft performance in strong wind conditions near vertical structures, with the goal of improving reliability and precision in aerial manipulation tasks conducted in close proximity to built environments. 

In summary, the key contributions of this paper are as follows:

\begin{itemize}
    \item A physics-based blade element model is integrated with a rotor-speed allocation strategy to more accurately capture wind-induced variations in rotor-generated forces and to mitigate aerodynamic disturbances at the actuation level.

    \item A neural network--based disturbance estimator is employed to predict and compensate for residual unmodeled aerodynamic effects, with analysis of modeling accuracy and data efficiency.

    \item A simulation environment is constructed to emulate strong horizontal and vertical wind conditions near a tall vertical wall.

    \item The proposed controller is evaluated through free-flight and wall-contact trajectory-tracking experiments, including comparative analysis of tracking accuracy and disturbance estimation performance with and without the blade element model.
\end{itemize}

\section{Modeling} 

\subsection{Aerial Manipulator Dynamics}

We consider a general dynamics model of an aerial manipulator~\cite{he2023image}:
\begin{equation}
    \mathbf{M} \dot{\mathbf{V}} + \mathbf{C}\mathbf{V}
    = \boldsymbol{\tau}_{\text{ext}} + \mathbf{G},
    \label{eq:lagrangian_model}
\end{equation}
where 
\begin{align}
    \mathbf{V} &= 
        \begin{bmatrix}
            \mathbf{v} \\
            \boldsymbol{\omega}
        \end{bmatrix}, \quad
    \mathbf{M}
    =
    \begin{bmatrix}
    m\,\mathbf{I}_{3\times 3} & \mathbf{0}_{3\times 3} \\
    \mathbf{0}_{3\times 3} & \mathbf{J}
    \end{bmatrix},
    \quad
    \mathbf{C}
    =
    \begin{bmatrix}
    m[\boldsymbol{\omega}]_\times & \mathbf{0}_{3\times 3} \\
    \mathbf{0}_{3\times 3} & [\boldsymbol{\omega}]_\times \mathbf{J}
    \end{bmatrix},
    \label{eq:mass_coriolis_gravity_correct} \quad
    \mathbf{G} = m\,
        \begin{bmatrix}
        \mathbf{R} \\
        \mathbf{0}_{3\times 3}
        \end{bmatrix}
        \mathbf{g}.
    \nonumber
\end{align}
Here, $[\cdot]_\times$ denotes the standard skew-symmetric matrix.

In the context of flight under strong wind conditions and proximity-induced
aerodynamic effects, the external wrench acting on the vehicle comprises two
components: the aerodynamic wrench $\boldsymbol{\tau}_a$ and the contact wrench
$\boldsymbol{\tau}_c$, 
\begin{equation}
    \boldsymbol{\tau}_{\text{ext}} = \boldsymbol{\tau}_a + \boldsymbol{\tau}_c,
    \label{eq:external_torque_decomposition}
\end{equation}
In this work, we focus on accurately modeling the aerodynamic component of the external wrench and treat the contact wrench as an unmodeled disturbance.

The aerodynamic wrench \( \boldsymbol{\tau}_a \in \mathbb{R}^6 \), consisting of the net force \( \mathbf{F}_a \) and moment \( \mathbf{T}_a \) acting on the vehicle, is given by:
\begin{equation}
    \boldsymbol{\tau}_a =
    \begin{bmatrix}
        \mathbf{F}_a \\
        \mathbf{T}_a
    \end{bmatrix},
    \label{eq:tau_a_definition}
\end{equation}
In this paper, we consider a quadrotor platform, and the total aerodynamic force and moment are given by
\begin{align}\label{eq:total_force} 
    \mathbf{F}_a = \sum_{i=1}^{4} \mathbf{F}_i, \quad
    \mathbf{T}_a = \sum_{i=1}^{4} \left( \mathbf{r}_i \times \mathbf{F}_i + \mathbf{T}_i \right)
\end{align}
For a standard quadrotor in a cross configuration, the rotor positions are:
\[
\mathbf{r}_1 = \begin{bmatrix} d_h \\ 0 \\ d_v \end{bmatrix}, \quad
\mathbf{r}_2 = \begin{bmatrix} 0 \\ d_h \\ d_v \end{bmatrix}, \quad
\mathbf{r}_3 = \begin{bmatrix} -d_h \\ 0 \\ d_v \end{bmatrix}, \quad
\mathbf{r}_4 = \begin{bmatrix} 0 \\ -d_h \\ d_v \end{bmatrix}.
\]

Notice that the aerodynamic force \( \mathbf{F}_i \) produced by each rotor includes not only axial thrust but also lateral components due to wind and induced flow interactions, while most of existing work on aerial manipulation assumes that rotors generate purely thrust-aligned forces.

\subsection{Rotor Force and Moment Modeling}

To model the aerodynamic thrust, drag, and axial torque generated by each rotor under varying wind and vehicle motion, we adopt Blade Element Momentum Theory (BEMT). BEMT is a well-established framework for propeller aerodynamics \cite{gill2017propeller}. It combines blade element theory, which evaluates the sectional lift and drag along the blade span from the local flow conditions, with momentum theory, which imposes a consistency condition on the induced velocity through the rotor disk. The induced velocity is obtained by solving a root-finding problem that matches the thrust predicted by blade element forces with that required by momentum theory. Once the induced flow is determined, the sectional forces and moments are integrated over azimuth and blade span to obtain the rotor-level total aerodynamic force and torque.

From BEMT, the aerodynamic force and torque generated by rotor~\(i\) are
modeled as
\begin{align}
    \mathbf{F}_i &= \mathbf{F}_{\text{BEMT}}\!\left(
        n_i,\;
        \mathbf{v}_{\text{wind},i},\;
        \mathbf{v}_{\text{prop},i}
        \;;\;
        c_{\ell,1},\;
        c_{\ell,2},\;
        c_d,\;
        \alpha_0,\;
        \theta(\cdot),\;
        c(\cdot)
    \right),
    \label{eq:Fi_model} \\
    \mathbf{T}_i &= \mathbf{T}_{\text{BEMT}}\!\left(
        n_i,\;
        \mathbf{v}_{\text{wind},i},\;
        \mathbf{v}_{\text{prop},i}
        \;;\;
        c_{\ell,1},\;
        c_{\ell,2},\;
        c_d,\;
        \alpha_0,\;
        \theta(\cdot),\;
        c(\cdot)
    \right).
    \label{eq:Ti_model}
\end{align}
where the variables and parameters appearing are summarized in Table~\ref{tab:symbols}. More details of BEMT can be found in~\cite{gill2017propeller}, and the numerical procedure used to evaluate $\mathbf{F}_{\mathrm{BEMT}}$ and $\mathbf{T}_{\mathrm{BEMT}}$ is provided in Appendix~\ref{app:bemt}. The blade geometry, defined by the pitch distribution $\theta(\cdot)$ and chord distribution $c(\cdot)$, can be directly measured, while the aerodynamic parameters $c_{\ell,1}$, $c_{\ell,2}$, $c_d$, and $\alpha_0$ characterize the airfoil lift and drag behavior. Together with $n_i$, $\mathbf{v}_{\text{wind},i}$, and $\mathbf{v}_{\text{prop},i}$, these quantities determine the rotor forces and moments under arbitrary wind conditions. Lateral aerodynamic moments are neglected; only the forces and the yaw moment along the rotor axis are considered.

\begin{figure}[htbp]
    \centering
    \includegraphics[width=0.7\linewidth]{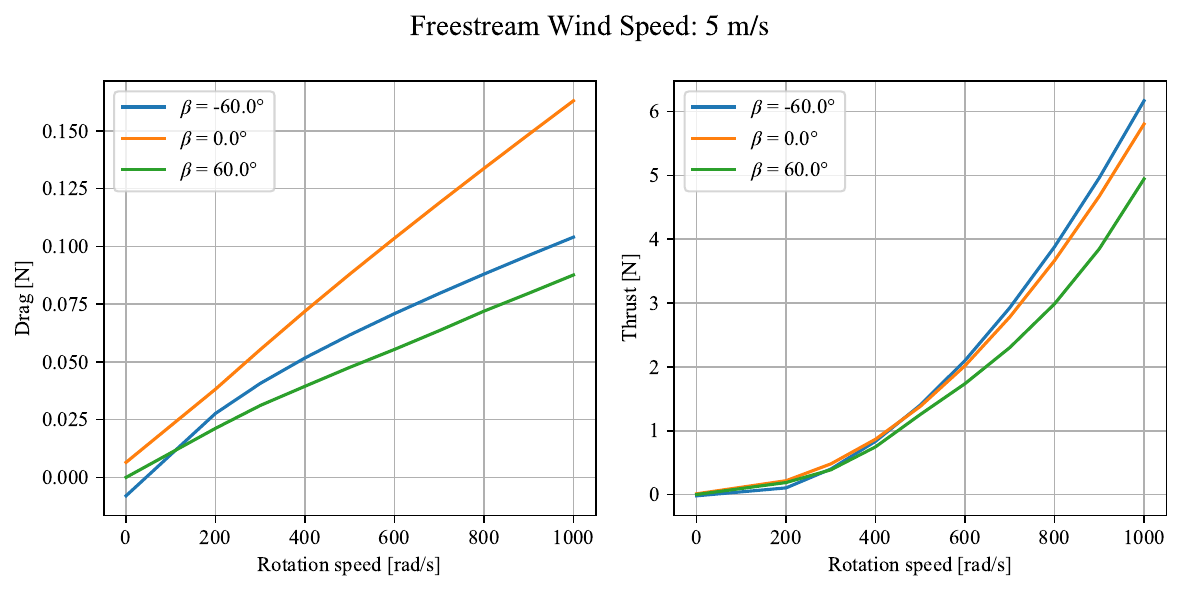}
    \caption{Rotor thrust and drag characteristics of the APC 8$\times$6 propeller~\cite{apc8x6} under varying pitch angles $\beta$ at a wind speed of 5\,m/s, evaluated across different rotational speeds.}
    \label{fig:bemt_combined}
\end{figure}

Figure~\ref{fig:bemt_combined} illustrates an example of how the BEMT model captures wind influences on rotor thrust and drag. The aerodynamic drag generated by the rotor increases with rotational speed, while the thrust is strongly influenced by the inflow angle~$\beta$, defined as the angle between the freestream wind direction and the rotor disk plane. The left subfigure illustrates the aerodynamic drag force acting parallel to the rotor disk plane, which grows monotonically with increasing rotational speed. The right subfigure shows the thrust variation across different pitch angles, highlighting how the thrust–speed relationship deviates from the nominal static-hover condition. These results emphasize that under varying wind flow angles, the propeller's rotation–thrust characteristics can differ significantly from those observed in static conditions, indicating the importance of incorporating aerodynamic effects in control and estimation.

Since BEMT reveals the underlying relationship between freestream conditions, propeller states, and the resulting aerodynamic wrench, the variation in aerodynamic forces and moments can be explained by combining BEMT with a spatially non-uniform flow field. In Section~\ref{sec:simulation_environment}, we demonstrate how BEMT can be used to model disturbance variations near a wall (i.e., the wall effect) by incorporating a flow field distribution.

\subsection{Motor Dynamics Modeling}
To account for the finite response time of the motor and electronic speed
controller, the rotor dynamics are modeled using a first-order approximation,
following the approach in \cite{eschmann2024data}. The actual rotor speed $n_i$
tracks the commanded value $n_{i,\text{cmd}}$ according to
\begin{equation}
    \dot{n}_i = \frac{1}{T_m}\left(n_{i,\text{cmd}} - n_i\right).
\end{equation}

This model captures the inherent actuation delay in rotor speed response and yields a more realistic coupling between control allocation and the aerodynamic wrench generated by the propellers. 

\section{Physics-Infused Learning for Disturbance Compensation}
\label{sec:method}

This section introduces the modeling and estimation framework used to characterize the overall disturbance wrench and to incorporate this information into a disturbance-aware controller.
We begin with the general rigid-body control model
\begin{equation}
    \mathbf{M}\dot{\mathbf{V}} + \mathbf{C}\mathbf{V}
    = \boldsymbol{u} + \boldsymbol{\tau}_d + \mathbf{G},
    \label{eq:control_law}
\end{equation}
where $\boldsymbol{u}$ is the control input generated by a geometric tracking controller~\cite{lee2010geometric}, and $\boldsymbol{\tau}_d$ denotes the net disturbance wrench, including aerodynamic uncertainty and contact-induced forces. Our objective is to construct a disturbance-compensation term that can be incorporated into the control law while remaining compatible with the available onboard sensing.

Because typical onboard sensors do not provide sufficiently reliable information to reconstruct aerodynamic torques, the disturbance observer operates at the force level. Let $\bar{\mathbf{F}}_a \in \mathbb{R}^3$ denote the nominal aerodynamic force predicted by the BEMT model, and let $\mathbf{F}_{\mathrm{phys}} = [\bar{F}_x,\; \bar{F}_y,\; 0]^\top$ denote its lateral (non-axial) component. The disturbance wrench used in the controller is then taken to be
\begin{equation}
\hat{\boldsymbol{\tau}}_d
=
\begin{bmatrix}
\hat{\mathbf{F}}_d \\
\mathbf{0}
\end{bmatrix},
\label{eq:disturbance_wrench_definition}
\end{equation}
where the estimated disturbance force is decomposed into three complementary contributions,
\begin{equation}
\hat{\mathbf{F}}_d
=
\mathbf{F}_{\mathrm{phys}}
+
\mathbf{F}_{\mathrm{res}}
+
\hat{\boldsymbol{\epsilon}} .
\label{eq:three_part_disturbance_summary}
\end{equation}
Here, $\mathbf{F}_{\mathrm{phys}}$ provides a physics-based nominal prediction, $\mathbf{F}_{\mathrm{res}}$ represents a learned residual component capturing unmodeled aerodynamic effects and $\hat{\boldsymbol{\epsilon}}$ is an adaptive correction that compensates for remaining, time-varying discrepancies.
A full definition of the composite tracking error $\mathbf{s}$ is given in Appendix~\ref{app:s}. 

The remainder of this section explains how each term in \eqref{eq:three_part_disturbance_summary} is constructed from data and integrated into the controller.

\subsection{Physics-based Learning of Aerodynamic Parameters}
\label{subsec:aero_parameter_identification}
To integrate the BEMT model into the controller, we first learn the blade-level aerodynamic coefficients \(c_{\ell,1}\), \(c_{\ell,2}\), \(c_d\), and \(\alpha_0\) using free-flight data. Because all four rotors employ identical propellers, a shared set of coefficients is assumed. These parameters are collected into the vector
\[
x_a = (c_{\ell,1},\; c_{\ell,2},\; c_d,\; \alpha_0).
\]
The learning procedure is physics-based, as it retains the structure of the BEMT model and estimates only the aerodynamic parameters that govern sectional lift and drag, rather than training a generic mapping from data. Because torque computation requires angular acceleration, which may not be reliably available, the learning procedure relies solely on the force component.

In the free-flight data collection, the contact wrench acting on the vehicle is zero. Thus, substituting \eqref{eq:external_torque_decomposition} and \eqref{eq:tau_a_definition}
into \eqref{eq:lagrangian_model} yields the net aerodynamic force acting on the vehicle:
\begin{equation}
    \mathbf{F} =  
    m \dot{\mathbf{v}} 
    + m [\boldsymbol{\omega}]_\times \mathbf{v}
    + m\, \mathbf{R} \mathbf{g}.
    \label{eq:aeroforce}
\end{equation}

From onboard IMU and motion-capture measurements, an estimation of the same force can be reconstructed as
\begin{equation}
    \mathbf{F}_{\text{sensed}} =  
    m \dot{\mathbf{v}}_{\text{sensed}} 
    + m [\boldsymbol{\omega}_{\text{sensed}}]_\times \mathbf{v}_{\text{sensed}}
    + m\, \mathbf{R} \mathbf{g}.
    \label{eq:aeroforce_sensed}
\end{equation}

Since individual rotor forces are not directly measurable, the learning is performed at the level of the net aerodynamic force. The nominal wrench predicted by the BEMT model is defined as
\begin{equation}
    \bar{\boldsymbol{\tau}}_a = 
    \begin{bmatrix}
        \bar{\mathbf{F}}_a \\
        \bar{\mathbf{T}}_a
    \end{bmatrix},
    \label{eq:tau_a_bemt}
\end{equation}
where, following \eqref{eq:total_force} and \eqref{eq:Fi_model}, the predicted net force is given by
\begin{equation}
    \bar{\mathbf{F}}_a(x_a) = 
    \sum_{i=1}^{4} 
    \mathbf{F}_\text{BEMT}\!\left(
        n_i,\;
        \mathbf{v}_{\text{wind},i},\;
        \mathbf{v}_{\text{prop},i}
        \;;\;
        x_a,\;
        \theta(\cdot),\;
        c(\cdot)
    \right),
    \label{eq:F_bar_def}
\end{equation}
with the rotor-hub translational velocity, the local freestream velocity, and the measured rotational speeds obtained from flight data, and with the blade-shape functions \(\theta(\cdot)\) and \(c(\cdot)\) specified by the propeller geometry. The local wind velocity can be estimated using onboard sensors in combination with wind flow estimation techniques, which is beyond the scope of this paper.

The aerodynamic parameter vector \(x_a\) is then learned by minimizing the mismatch between the BEMT-predicted force and the sensor-reconstructed force:
\begin{equation}
    x_a^{*} = 
    \arg\min_{x_a}
    \left\|
        \bar{\mathbf{F}}_a(x_a) - \mathbf{F}_{\text{sensed}}
    \right\|_2^{2}.
    \label{eq:param_identification}
\end{equation}

The resulting parameters define a physics-informed nominal model, and the rotor force and torque maps used by the controller are constructed directly from this learned model. Because each BEMT evaluation requires solving a root-finding problem, computing these maps in real time is too costly for onboard processors. Instead, the force and moment values are precomputed over the operating range and stored in a lookup table, allowing fast evaluation during flight.

\subsection{BEMT-based Propeller Rotational Speed Allocation}
\label{subsec:bemt_speed_allocation}
Equipped with the nominal BEMT model and the learned aerodynamic parameters, we can allocate rotor speeds with improved accuracy. We assume that the rotational speed \(n_i\) of each rotor is directly controllable. Given the desired total thrust and moments acting on the vehicle, the corresponding thrusts allocated to the individual propellers satisfy the following relation:

\begin{equation}
\begin{bmatrix}
F_z \\
M_x \\
M_y \\
M_z
\end{bmatrix}
=
\begin{bmatrix}
1 & 1 & 1 & 1 \\
0 & -d_h & 0 & d_h \\
d_h & 0 & -d_h & 0 \\
-c_{\tau f} & c_{\tau f} & -c_{\tau f} & c_{\tau f}
\end{bmatrix}
\begin{bmatrix}
f_1 \\
f_2 \\
f_3 \\
f_4
\end{bmatrix}
\end{equation}
where \( f_i \) denotes the allocated thrust generated by rotor \(i\) for \(i = 1,2,3,4\); \( F_z \) is the total thrust command along the body \(z\)-axis; \( M_x \), \( M_y \), and \( M_z \) are the body-frame moment command about the \(x\)-, \(y\)-, and \(z\)-axes, respectively; and \( c_{\tau f} \) is the torque coefficient relating rotor thrust to the reactive (yaw) torque.

The commanded rotational speed \(n_{i,\text{cmd}}\) is obtained by solving
for the rotor speed that produces the allocated thrust \(f_i\):
\begin{equation}
\begin{aligned}
    n_{i,\text{cmd}}^{*}
    &= \arg\min_{n} \; (n - n_i)^2 \\
    \text{s.t.}\;\;
    f_i
    &= 
    \left[
        \mathbf{F}_\text{BEMT}\!\left(
            n,\;
            \mathbf{v}_{\text{wind},i},\;
            \mathbf{v}_{\text{prop},i}
            \;;\;
            x_a,\;
            \theta(\cdot),\;
            c(\cdot)
        \right)
    \right]_z ,
\end{aligned}
\label{eq:thrust_matching}
\end{equation}
for each rotor \(i = 1,\ldots,4\). Note that the mapping \(n \mapsto \left[\mathbf{F}_\text{BEMT}(n,\cdot)\right]_z\) is generally not injective, and multiple rotor speeds may satisfy the thrust constraint. The optimization in \eqref{eq:thrust_matching} therefore selects, among all feasible solutions, the rotational speed closest to the measured rotational speed \(n_i\), ensuring continuity and preventing nonphysical jumps in the commanded motor input. The resulting horizontal force and yaw moment mismatch are treated as external disturbances and delegated to the upstream controller. Compared to the traditional quadratic thrust model, this approach reduces thrust mismatch and mitigates the unintended lateral moment applied to the vehicle’s center of mass.

We now consider the physics-based disturbance term used in the decomposition introduced earlier. The nominal aerodynamic force predicted by the BEMT model is
\[
\bar{\mathbf{F}}_a = (\bar{F}_x,\; \bar{F}_y,\; \bar{F}_z)^\top,
\]
where \(\bar{F}_z\) corresponds to the thrust direction. Because the allocation procedure selects rotor speeds that reproduce the desired thrust, the quantity \(\bar{F}_z\) largely reflects the commanded thrust rather than an external
aerodynamic disturbance. In contrast, the lateral components \((\bar{F}_x,\bar{F}_y)\) capture aerodynamic loading that is not explicitly regulated by the thrust input. Motivated by this distinction, we define the
lateral nominal aerodynamic force
\[
\bar{\mathbf{F}}_{a,\mathrm{lat}}
= (\bar{F}_x,\; \bar{F}_y,\; 0)^\top,
\]
and use it as the physics-based disturbance contribution in \eqref{eq:three_part_disturbance_summary},
\begin{equation}
\mathbf{F}_{\mathrm{phys}} = \bar{\mathbf{F}}_{a,\mathrm{lat}}.
\label{eq:Fphys_def}
\end{equation}

Although motor dynamics introduce small deviations between commanded and realized thrust, these axial mismatches cannot be cleanly decoupled from the control input and are therefore not included as disturbances.

\subsection{Physics-infused Residual Learning}

Although the physics-based learning stage adapts the BEMT model to data, it is limited to the blade-level parameters and therefore cannot capture all aerodynamic variations that arise during flight. To provide the necessary flexibility, we augment the nominal BEMT model with a lightweight neural network that learns the remaining unmodeled aerodynamic force components. This yields a physics-infused disturbance estimation framework in which the learned BEMT model provides the nominal aerodynamic wrench, while the neural residual model compensates for effects not captured by the physics-based parameterization, including sensor noise, biases, and unmodeled aerodynamic phenomena.

A multi-head residual network is trained to model the mismatch between
the BEMT-predicted forces and the measured forces. The network minimizes
the loss
\[
\mathcal{L} = 
\sum_{i=1}^{N} 
\left\| 
\mathbf{F}_{\text{sensed}}^{(i)} 
- 
\bar{\mathbf{F}}_a^{(i)} 
- 
\boldsymbol{\phi}\!\left(
    \mathbf{x}_\phi^{(i)}
\right)
\right\|^{2}.
\]
where
\begin{itemize}
    \item $\boldsymbol{\phi}$ denotes the neural network that predicts the
    residual aerodynamic force,
    \item $\mathbf{x}_\phi^{(i)}$ is the input for sample $i$, consisting of  
    \begin{itemize}
        \item the vehicle twist (6-d),  
        \item the pose represented as a quaternion (4-d),  
        \item the rotational speed of each rotor (1-d per rotor), and  
        \item the local freestream wind velocity at each rotor hub (3-d per rotor),   
    \end{itemize}
    \item $\mathbf{F}_{\text{sensed}}^{(i)} $ is the sensed net aerodynamic force for sample $i$,
    \item $\bar{\mathbf{F}}_a^{(i)}$ is the net aerodynamic force predicted by the nominal BEMT model for sample $i$,
    \item $N$ is the total number of training samples.
\end{itemize}
Thus, the input dimension is $n_\phi = 26$.

The neural network $\boldsymbol{\phi}$ adopts a multi-head architecture with a shared backbone and separate output heads, each responsible for predicting one dimension of the residual force. Training is performed using the Adam optimizer with a learning rate of $5\text{e-}4$ for 200 epochs. Spectral normalization is applied to all layers to ensure bounded outputs. The objective minimizes the discrepancy between the predicted and measured disturbance force while enforcing Lipschitz continuity as a regularizing constraint. The network hyperparameters are summarized in Table~\ref{tab:physics_infused_hyperparams}.

\begin{table}[H]
    \centering
    \begin{threeparttable}
    \caption{Hyperparameters of the residual network}
    \label{tab:physics_infused_hyperparams}
    \begin{tabular}{ll}
        \toprule
        \textbf{Component} & \textbf{Value / Description} \\
        \midrule
        Shared network architecture & 26 $\rightarrow$ 64 $\rightarrow$ 32 (ReLU) \\
        Each head architecture & 32 $\rightarrow$ 32 $\rightarrow$ 3 (ReLU) \\
        Batch size ($B$) & 256 \\
        Learning rate & 0.0005 \\
        \bottomrule
    \end{tabular}
    \end{threeparttable}
\end{table}

Similar to the dataset used for physics-based learning, the training dataset is generated by flying randomized trajectories in open space under a range of wind conditions. These conditions are summarized along with the flight configuration in Table~\ref{tab:training_wind_conditions}.

\begin{table}[H]
\centering
\begin{threeparttable}
\caption{Training data generation configuration}
\label{tab:training_wind_conditions}
\renewcommand{\arraystretch}{1.15}
\begin{tabular}{ll}
\toprule
\textbf{Parameter} & \textbf{Specification} \\
\midrule
Horizontal wind set & \(\{-3,\,-1,\,0\}\)\,m/s \\
Vertical wind set   & \(\{-5,\,-3,\,0,\,3,\,5\}\)\,m/s \\
Trajectory profile  & Randomized 3D trajectory in open space \\
Flight duration per trial & 60\,s \\
\bottomrule
\end{tabular}

\begin{tablenotes}[flushleft]
\footnotesize
\item Negative horizontal values indicate headwind, while vertical values indicate downdraft 
(negative) or updraft (positive). All combinations of the horizontal and vertical wind values are used.
\end{tablenotes}

\end{threeparttable}
\end{table}

During each flight, the vehicle's pose, linear and angular velocities, linear acceleration, propeller rotational speeds, and the local ambient wind velocity at the rotor hubs are recorded to construct the training dataset. Wind conditions are generated using discrete speed intervals ranging from 0 to 3~m/s in the horizontal direction and from $-5$ to 5~m/s in the vertical direction. Because the BEMT model computes the aerodynamic wrench independently of rotor location, near-wall flight data are not required for training to capture near-wall effects.

Finally, the network output is taken to represent the residual disturbance term in \eqref{eq:three_part_disturbance_summary},
i.e.,
\begin{equation}
\mathbf{F}_{\mathrm{res}} = \boldsymbol{\phi}(\mathbf{x}_\phi),
\label{eq:Fres_def}
\end{equation}
Thus, the neural model provides a data-driven correction to the nominal aerodynamic force, complementing the physics-based term $\mathbf{F}_{\mathrm{phys}}$ and the online adaptive component $\hat{\boldsymbol{\epsilon}}$. 

\subsection{Adaptive Residual Disturbance Observer}

To compensate for the time-varying and unmodeled effects that remain after the physics-based and residual-learning components, an online estimator updates the adaptive term \( \hat{\boldsymbol{\epsilon}} \) in \eqref{eq:three_part_disturbance_summary}. The evolution of the estimator is specified by the adaptation law and covariance update:
\begin{equation}
\dot{\hat{\boldsymbol{\epsilon}}}
=
-\lambda \hat{\boldsymbol{\epsilon}}
- \mathbf{P}\mathbf{R}^{-1} \left(\hat{\boldsymbol{\epsilon}} - \boldsymbol{y}\right)
+ \mathbf{P}\mathbf{H}\mathbf{s},
\label{eq:adaptation_law0}
\end{equation}
\begin{equation}
\dot{\mathbf{P}}
=
-2\lambda \mathbf{P}
+ \mathbf{Q}
- \mathbf{P}\mathbf{R}^{-1}\mathbf{P},
\label{eq:covariance_update0}
\end{equation}
where \(\mathbf{H} = \begin{bmatrix}\mathbf{I}_{3\times 3} & \mathbf{0}_{3\times 3}\end{bmatrix} \)
selects the translational components of the tracking error, \(\mathbf{P} \in \mathbb{R}^{3\times 3}\) is the covariance matrix, \(\mathbf{R} \in \mathbb{R}^{3\times 3}\) and \(\mathbf{Q} \in \mathbb{R}^{3\times 3}\) are symmetric positive-definite gain matrices, and \(\lambda > 0\) is a regularization coefficient. The term \(\boldsymbol{y}\) appearing in the adaptation law represents the measured residual force and is defined as
\begin{equation}
\boldsymbol{y}
=
\mathbf{F}_{\text{sensed}}
-
\left(\bar{\mathbf{F}}_a + \boldsymbol{\phi}(\mathbf{x}_\phi)\right),
\label{eq:measured_residual_wrench}
\end{equation}
capturing the discrepancy between the sensed aerodynamic force and the combined prediction of the nominal BEMT model and the neural network. 
Finally, combining the physics-based term in \eqref{eq:Fphys_def}, the learned residual term in \eqref{eq:Fres_def}, the adaptive update laws \eqref{eq:adaptation_law0}--\eqref{eq:measured_residual_wrench}, and the three-part decomposition \eqref{eq:three_part_disturbance_summary} yields the complete disturbance-compensation scheme employed in the controller.

\section{Simulation Environment Setup}
\label{sec:simulation_environment}

The simulation models a scenario in which a small quadrotor flies in close proximity to a tall building surface, subject to wind disturbances originating from horizontal, vertical, or combined directions. The vehicle is a lightweight, symmetric quadrotor, and its physical parameters are summarized in Table~\ref{tab:quadrotor_params}.

\begin{table}[htbp]
    \centering
    \caption{Quadrotor simulation parameters used in the proximity-to-wall wind disturbance scenario.}
    \label{tab:quadrotor_params}
    \begin{tabular}{ll}
        \toprule
        \textbf{Parameter} & \textbf{Value / Description} \\
        \midrule
        Total mass $m$ & 2.1395 kg (drone + battery) \\
        Rotor horizontal distance $d_h$ & 0.28 m (center to rotor center) \\
        Rotor vertical offset $d_v$ & 0.095 m (above CoG) \\
        Inertia matrix $\mathbf{J}$ & diag(0.0820, 0.0845, 0.1377) kg·m\textsuperscript{2} \\
        Number of rotors & 4 \\
        Rotor configuration & Symmetric cross, 90° apart \\
        Rotor directions & CCW–CW–CCW–CW (motors 1–4) \\
        Propeller type & APC 8$\times$6 \\
        Torque coefficient $c_{\tau f}$ & $8.004 \times 10^{-3}$ \\
        \bottomrule
    \end{tabular}
\end{table}

\subsection{Near-Wall Wind Environment Configuration}
\begin{figure}[H]
    \centering
    \includegraphics[width=0.6\linewidth]{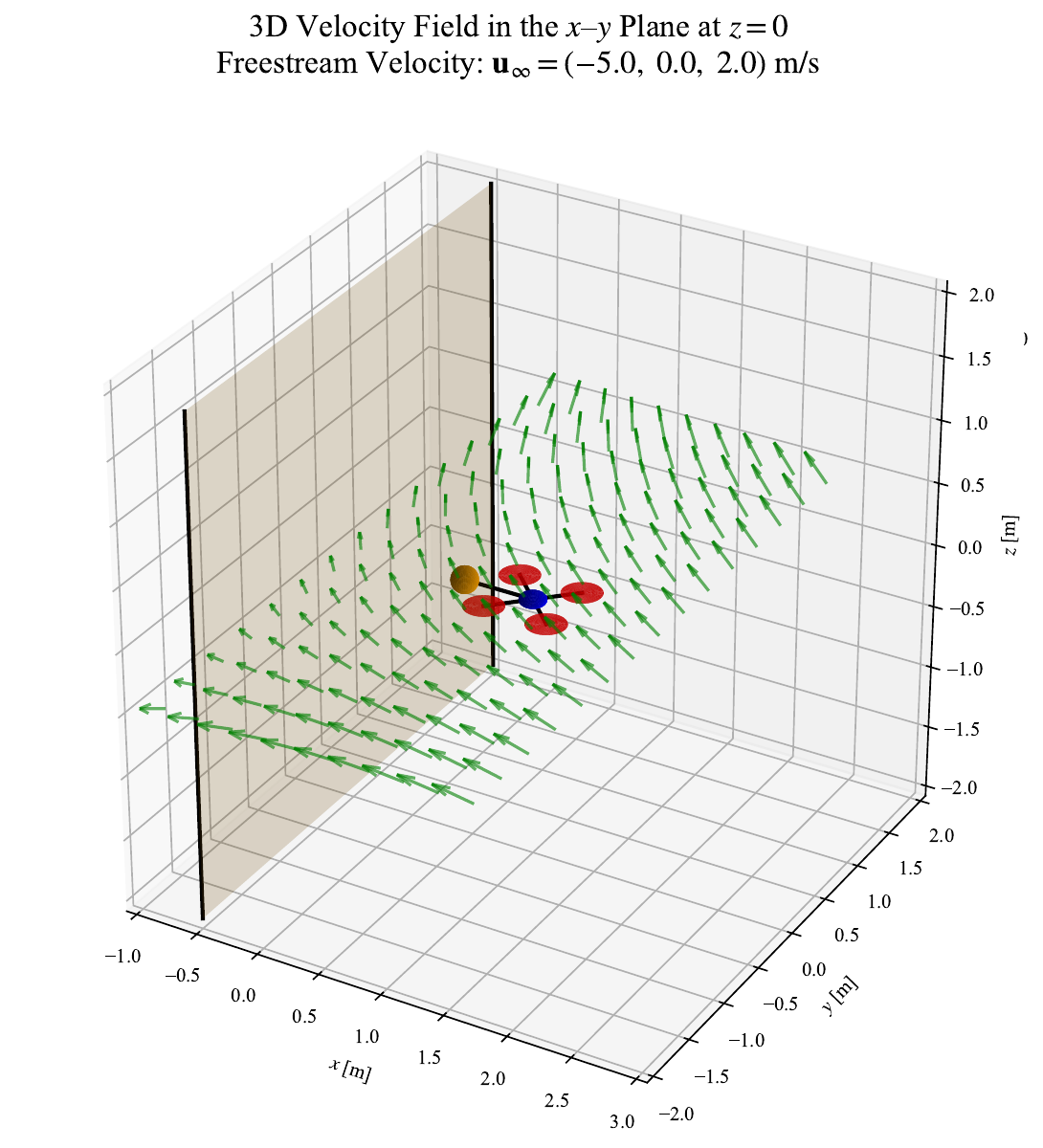}
    \caption{Simulated 3D velocity field in the $x$–$y$ plane at $z = 0$ for an oblique freestream wind}
    \label{fig:flow-field-3d}
\end{figure}

A tall, vertically aligned building surface can be idealized as having infinite height. When wind flows over such a large vertical structure, the aerodynamic behavior can be reasonably approximated using a two-dimensional potential flow model in the horizontal ($x$–$y$) plane. This approximation captures the horizontal redistribution of wind caused by the wall, while the vertical wind component (along the $z$-axis) is assumed to remain uniformly distributed throughout the domain, as illustrated in Figure~\ref{fig:flow-field-3d}. In our simulation, the wall is modeled as a 4\,m-long vertical panel centered at $(-0.5,\,0,\,0)$ and oriented to face the positive $x$-axis, representing a protruding element on a building facade.

The wall effect arises implicitly from how the nonuniform wind field alters the local flow experienced by each rotor. The wind velocity varies with the quadrotor's relative horizontal position with respect to the wall's centerline. As a result, the quadrotor may experience significantly different disturbance forces depending on its distance to the wall. Table~\ref{tab:disturbances_vs_wall} presents the simulated disturbance force and torque values when the quadrotor hovers at different distances from the wall along the vertical plane perpendicular to the wall and aligned with its centerline (i.e., the symmetry plane). 

\begin{table}[H]
\centering
\caption{Disturbance forces and torques at different wall distances during hover under a $-10~\mathrm{m/s}$ wind applied in the $x$-direction. Due to the relative positions of the rotors, lateral and non-uniform axial disturbance forces can induce a net disturbance torque on the quadrotor, even though BEMT does not account for lateral moments generated by individual rotors.}
\label{tab:disturbances_vs_wall}
\begin{tabular}{cccc}
\toprule
\textbf{Wall Distance (m)} & \textbf{$F_x$ (N)} & \textbf{$F_z$ (N)} & \textbf{$T_y$ (Nm)} \\
\midrule
0.5 & -0.396 & 0.429 & -0.006 \\
1.0 & -0.859 & 0.763 & 0.019 \\
5.0 & -2.778 & 1.249 & 0.110 \\
\bottomrule
\end{tabular}
\end{table}





The simulation environment employs a refined implementation of the BEMT model to generate the aerodynamic ground truth. This refined model uses a much denser numerical discretization (100 radial segments and 90 azimuthal segments) to produce high-fidelity force and torque predictions. In contrast, the parameter-learned BEMT model introduced in
Section~\ref{subsec:aero_parameter_identification} uses a coarser discretization (20 radial and 18 azimuthal segments). Although less refined, the learned model is sufficiently accurate for parameter estimation and for use within the controller. The intentional difference in discretization creates a realistic modeling gap between the nominal model and the simulation ground truth, reflecting the mismatch that typically exists between onboard aerodynamic models and true rotor aerodynamics.


\subsection{Wall-Contact Interaction Setup}
The quadrotor is equipped with a lightweight rigid rod mounted along the negative $x$-axis of the body frame, with a deformable spherical sponge attached at its distal end to serve as the contact tip. This end-effector setup enables the vehicle to make contact with the wall interaction tasks. When the sponge touches the wall, normal compression and tangential 
friction are introduced based on a linear deformation response and a Coulomb friction assumption. The relevant parameters are summarized in Table~\ref{tab:endeffector_params}.

\begin{table}[htbp]
    \centering
    \caption{End-effector geometry and contact model parameters}
    \label{tab:endeffector_params}
    \begin{tabular}{ll}
        \toprule
        \textbf{Parameter} & \textbf{Value} \\
        \midrule
        Rod length & 0.45 m \\
        Tip position in body frame & $[-0.45,\; 0,\; 0]^\top$ m \\
        Tip sponge radius & 0.1 m \\
        Linear spring constant & 10 N/m \\
        Coulomb friction coefficient  & 0.4 \\
        \bottomrule
    \end{tabular}
\end{table}

\subsection{Sensor Noise}
\begin{figure}[H]
    \centering
    \includegraphics[width=0.9\linewidth]{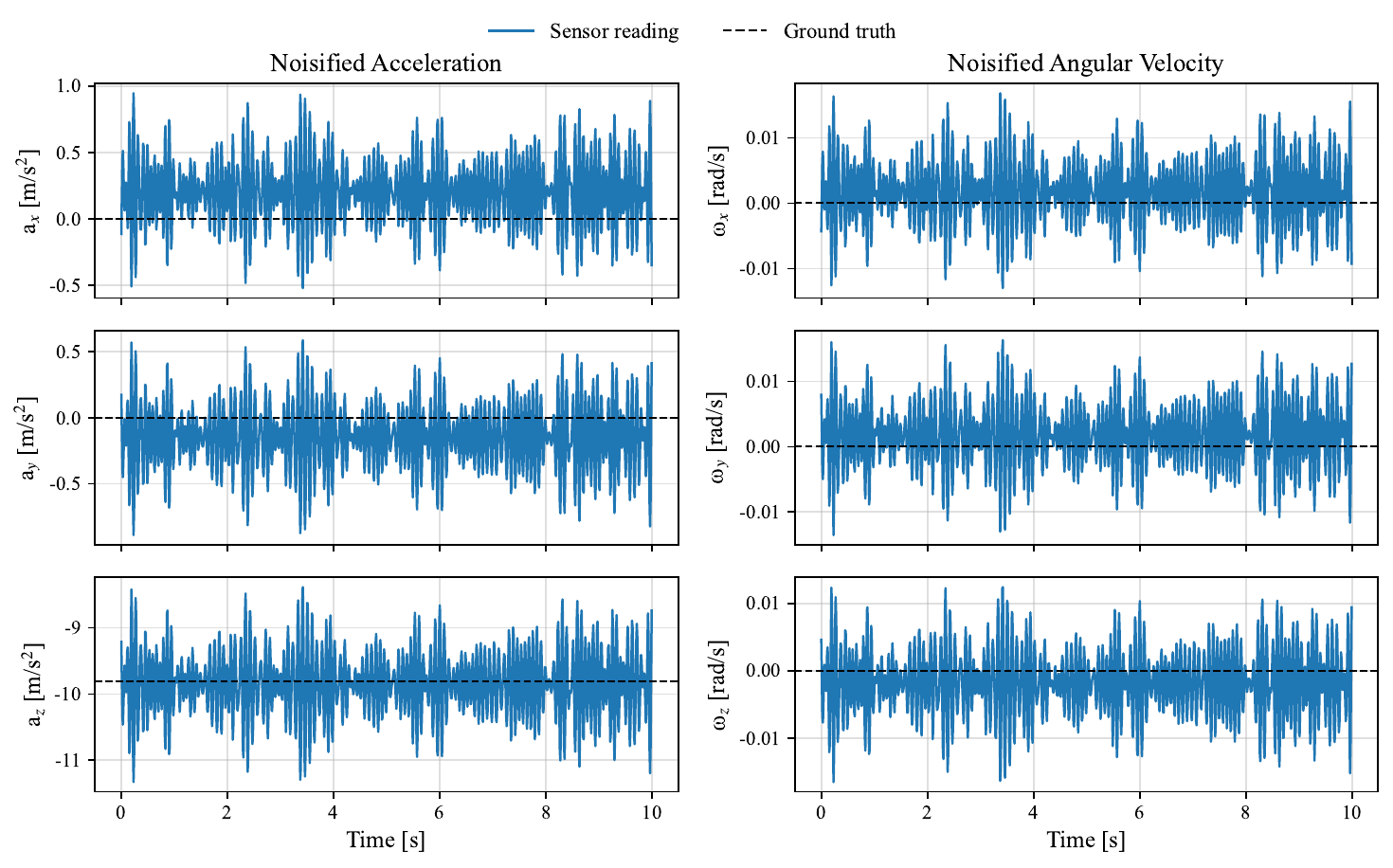}
    \caption{Noisified IMU acceleration and angular velocity measurements during a steady hover}
    \label{fig:imu}
\end{figure}

To simulate typical IMU noise, we adopt the specifications of the Epson G365 (M-G365PDC1/PDF1) IMU\cite{epsonG365}
. The IMU model includes a turn-on bias and a random walk that is modeled by white noise. An installation misalignment is also included to introduce additional sensor reading error. In addition, aerodynamically induced vibration is modeled as a series of banded harmonics with evenly distributed amplitudes and randomly assigned phases in the 10–15 Hz range, producing larger fluctuations in the simulated sensor readings. Figure~\ref{fig:imu} illustrates a representative instance of the simulated noisy IMU measurements obtained during a stationary hover.

\subsection{Task Trajectory}
\label{subsec:task_traj}
\begin{figure*}[htbp]
    \centering
    \begin{subfigure}[b]{0.48\textwidth}
        \centering
        \includegraphics[width=\linewidth]{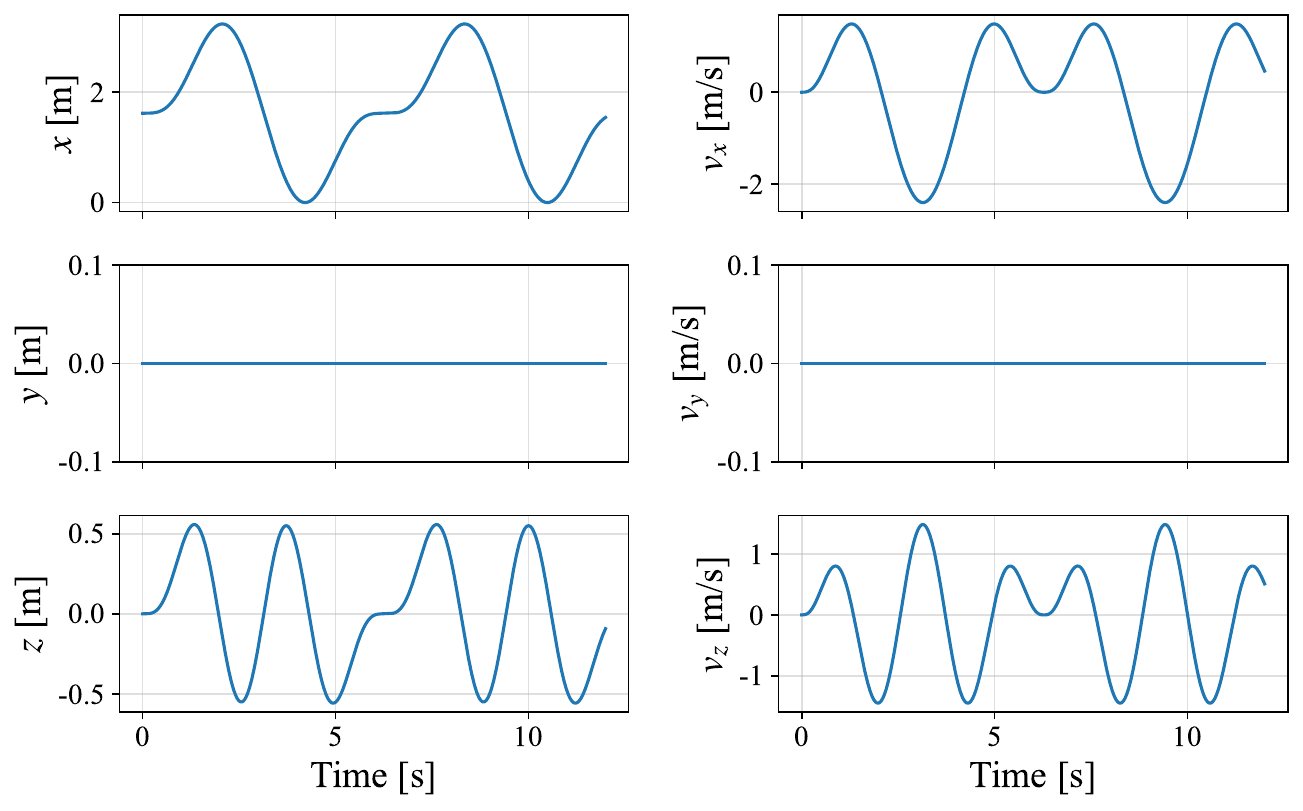}
        \caption{Figure-eight trajectory}
        \label{fig:traj_figure8}
    \end{subfigure}
    \hfill
    \begin{subfigure}[b]{0.48\textwidth}
        \centering
        \includegraphics[width=\linewidth]{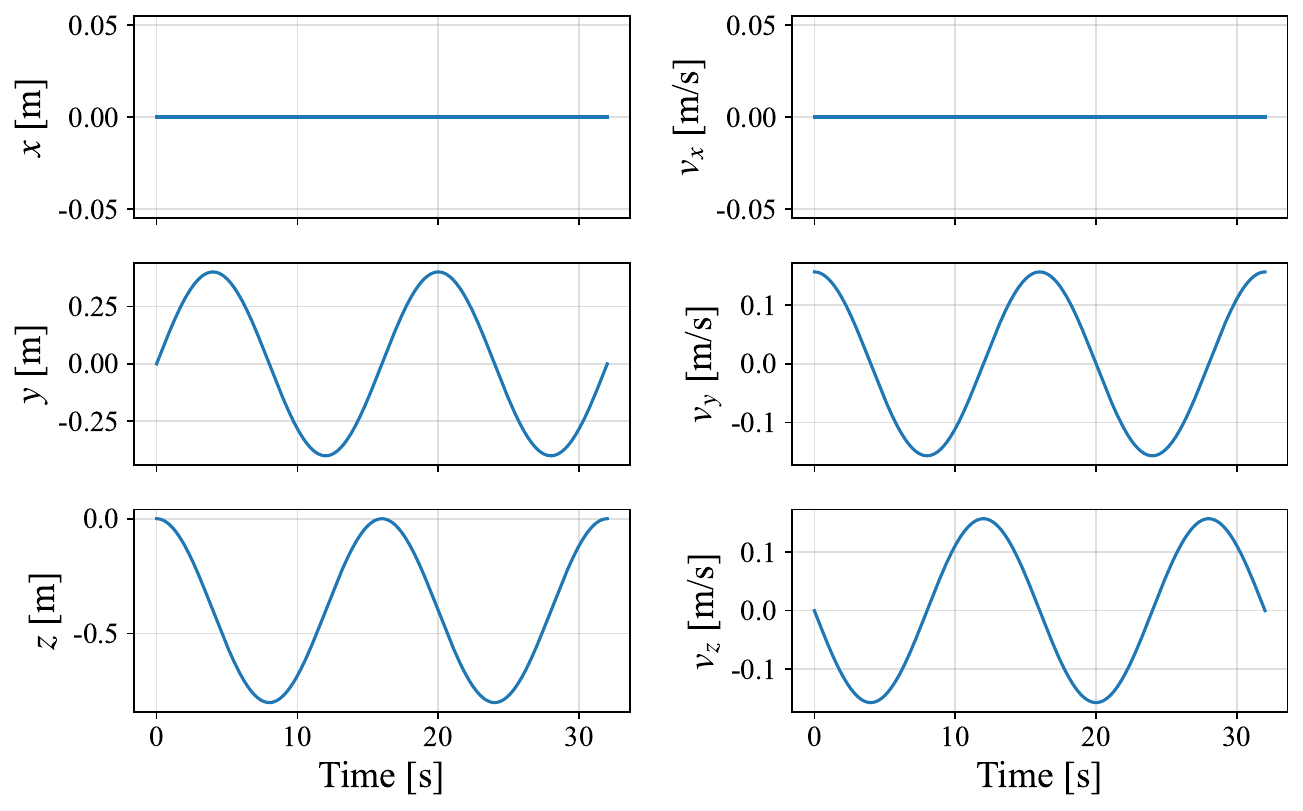}
        \caption{Circular trajectory}
        \label{fig:traj_circle}
    \end{subfigure}
    \caption{Simulation reference trajectories.}
    \label{fig:trajectories}
\end{figure*}

Two trajectory-tracking scenarios are considered. In the first, the drone follows 
a vertical figure-eight trajectory in a plane perpendicular to the wall to assess 
disturbance-rejection performance (Figure~\ref{fig:traj_figure8}). In the 
second, a vertical circular trajectory lying in a plane parallel to the wall is 
designed such that the end effector makes light, intentional contact at the 
desired location (Figure~\ref{fig:traj_circle}).

\section{Results and discussion}
\subsection{Aerodynamic Parameter Learning Results}

Table~\ref{tab:airfoil_param_comparison} summarizes the aerodynamic coefficients obtained from the free-flight parameter learning and compares them with the ground-truth values used in simulation. All four parameters characterize the airfoil-level behavior within the blade element theory formulation and therefore describe aerodynamics at the blade-element level rather than the net forces produced by the rotor as a whole. The learned coefficients exhibit a consistent tendency toward overestimation—\(c_{\ell,1}\) and \(c_{\ell,2}\) indicating stronger sectional lift response, \(c_d\) indicating a larger effective profile-drag scale, and \(\alpha_0\) indicating a slightly earlier transition into the post–stall regime.

\begin{table}[htbp]
    \centering
    \caption{Comparison of learned versus ground-truth aerodynamic coefficients.}
    \label{tab:airfoil_param_comparison}
    \renewcommand{\arraystretch}{1.15}
    \begin{tabular}{lcc}
        \toprule
        \textbf{Parameter} & \textbf{Learned Value} & \textbf{Ground Truth} \\
        \midrule
        $c_{\ell,1}$ & 5.460 & 5.300 \\
        $c_{\ell,2}$ & 2.060 & 1.700 \\
        $c_{d}$      & 3.007 & 1.800 \\
        $\alpha_0$ (rad) & 0.352 & 0.360 \\
        \bottomrule
    \end{tabular}
\end{table}

\begin{figure}[H]
    \centering
    \includegraphics[width=0.9\linewidth]{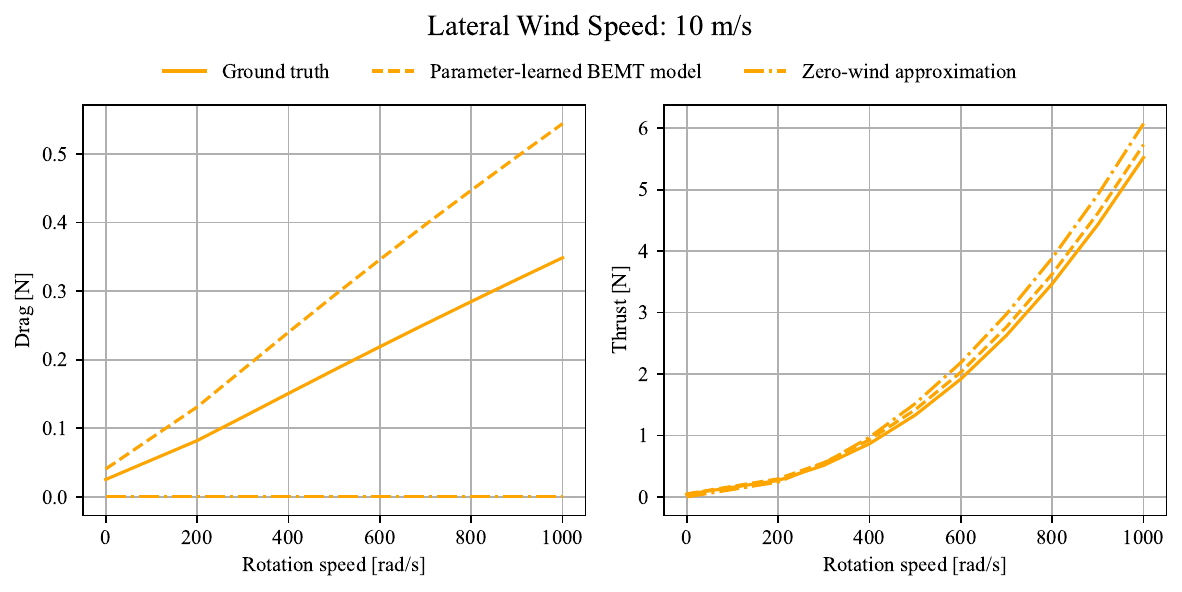}
    \caption{Rotor thrust and drag characteristics of the APC 8×6 propeller comparison among simulation groundtruth, parameter-learned BEMT model and zero-wind approximation.}
    \label{fig:bemt_table_compare}
\end{figure}

These biases manifest at the rotor level after integration over azimuth and blade span, as illustrated in 
Fig.~\ref{fig:bemt_table_compare}, where the parameter-learned BEMT model overpredicts both thrust and drag relative to the simulation ground truth under a 10~m/s lateral wind condition not seen during parameter learning. This distinction highlights that discrepancies in blade-element coefficients propagate through the integration process to produce differences in rotor-level force predictions.
Overall, the learned parameter set yields a nominal aerodynamic model of sufficient fidelity for controller deployment, while retaining a moderate degree of modeling error. This residual discrepancy is anticipated and is addressed by the additional compensation mechanisms introduced in this paper.

\subsection{Evaluation of Thrust Mapping in Wind}

Figure~\ref{fig:bemt_table_compare} shows that using a thrust map obtained under 
zero-wind conditions can introduce a noticeable variation in the predicted rotor 
    thrust once the vehicle encounters ambient airflow, in this case, an overestimation across the entire operational rotational speed range. This occurs because the zero-wind thrust mapping does not generalize. The same limitation also applies to the common approximation that thrust is strictly proportional to the square of the rotational speed. In contrast, the 
BEMT-based speed allocation accounts for wind effects and provides a more consistent thrust estimate across varying wind conditions.

\subsection{Residual Learning Performance}

A test dataset is collected by flying the figure-eight task defined in Subsection~\ref{subsec:task_traj} under a range of prescribed wind conditions. Table~\ref{tab:test_wind_conditions} summarizes the wind settings and flight specifications used for generating the test data. Compared to the training conditions in Table~\ref{tab:training_wind_conditions}, the test wind settings span a larger range.

\begin{table}[htbp]
\centering
\begin{threeparttable}
\caption{Test data generation configuration}
\label{tab:test_wind_conditions}
\renewcommand{\arraystretch}{1.15}
\begin{tabular}{ll}
\toprule
\textbf{Parameter} & \textbf{Specification} \\
\midrule
Horizontal wind set & \(\{-5,\,-2,\,0\}\)\,m/s \\
Vertical wind set   & \(\{-10,\,-4,\,-1,\,1,\,4,\,10\}\)\,m/s \\
Trajectory profile  & Figure-eight trajectory near wall \\
Flight duration per trial & 40\,s \\
\bottomrule
\end{tabular}

\begin{tablenotes}[flushleft]
\footnotesize
\item Negative horizontal values indicate headwind, while vertical values indicate downdraft 
(negative) or updraft (positive). All combinations of the horizontal and vertical wind values are used.
\end{tablenotes}

\end{threeparttable}
\end{table}

\begin{figure}[htbp]
    \centering
    \includegraphics[width=0.9\linewidth]{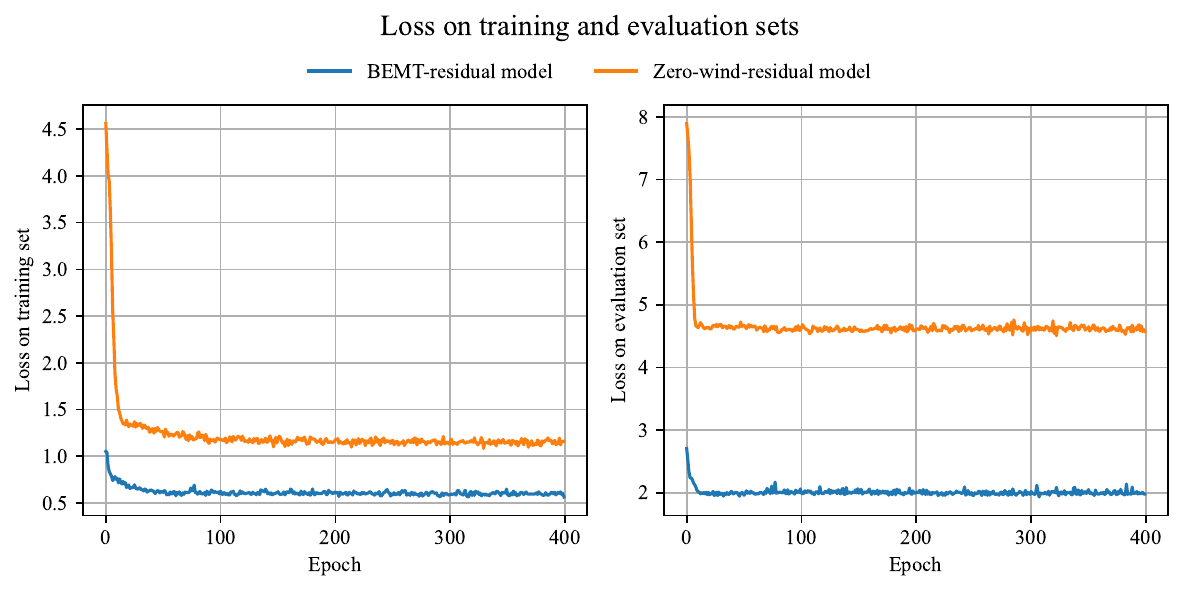}
    \caption{Comparison of training and evaluation losses for the BEMT-residual 
    model and the zero-wind-residual model. Incorporating the BEMT prior reduces the 
    overall loss on both datasets, indicating improved alignment between the 
    predicted and measured aerodynamic forces.}
    \label{fig:loss_comparison_bemt_vs_zerowind}
\end{figure}

Residual learning performance is evaluated on this dataset to assess its predictive accuracy relative to the nominal BEMT model. Although the dataset serves as a test set, its loss is monitored at every training epoch for comparison purposes only and does not influence optimization. Two residual-learning variants are considered: a BEMT-residual model, which learns the force mismatch with respect to the parameter-learned BEMT model, and a zero-wind-residual model, which learns the mismatch relative to the zero-wind approximation. The resulting training and test losses for both variants are reported in Fig.~\ref{fig:loss_comparison_bemt_vs_zerowind}. The comparison shows that the BEMT-residual model achieves consistently lower loss, indicating that the physics-based nominal model explains a larger portion of the aerodynamic effects and allows the residual network to concentrate on finer discrepancies that remain unmodeled.

\subsection{Tracking Performance}
In both the figure-eight and circular wall-contact scenarios, the trajectory-tracking performance of each controller is evaluated in terms of the RMS tracking error under constant wind disturbances, where wind is applied equally along the $x$- and $z$-directions at $-4$, $-8$, and $-12~\mathrm{m/s}$. These test conditions are intentionally chosen to exceed the wind magnitudes present in the training data in order to assess each controller’s ability to generalize to previously unseen wind intensities. The following three controller configurations are compared:
\begin{itemize}
    \item No disturbance compensator
    \item DAIML-based compensator, where DAIML refers to a domain adversarially invariant meta-learning strategy proposed in~\cite{o2022neural}
    \item Proposed learned BEMT-based compensator with rotor-speed allocation
\end{itemize}
DAIML learns a disturbance model that is invariant to domain shifts by aligning feature distributions across different wind conditions. Its representation is trained purely from data, without incorporating aerodynamic structure or any model-based priors.
In contrast, the proposed approach integrates a model-based nominal component with residual learning. This hybrid structure separates predictable aerodynamic effects from unmodeled phenomena, leading to improved robustness under strong and previously unseen wind conditions.

In the figure-eight scenario (Table~\ref{tab:tracking_comparison}), only the position tracking errors in the $x$- and $z$-directions are considered, as the trajectory lies in the symmetry plane of the wall and the wind has no component in the $y$-direction; consequently, the $y$-direction RMS error is negligible and omitted from the analysis. The configuration without a disturbance compensator maintains acceptable tracking performance under a $-4~\mathrm{m/s}$ wind. However, at $-8~\mathrm{m/s}$, wind-induced disturbances lead to noticeable bias in the negative $x$- and $z$-directions, and at $-12~\mathrm{m/s}$ the controller is unable to follow the trajectory.
The DAIML configuration achieves the smallest horizontal tracking error at $-4$ and $-8~\mathrm{m/s}$, remaining accurate and stable within the wind range encountered during training. In contrast, the $-12~\mathrm{m/s}$ case lies well outside the training distribution, and DAIML does not generalize effectively: although the vehicle remains stable, its tracking errors increase notably in both axes.
This behavior is consistent with the disturbance–wall interaction shown in Table~\ref{tab:disturbances_vs_wall}, where the horizontal disturbance force induced by wind decreases as the vehicle approaches the wall. Consequently, when flying farther from the wall, a larger cross-track error is expected.

The proposed method demonstrates tracking performance comparable to the DAIML configuration at $-4$ and $-8~\mathrm{m/s}$, with slightly smaller vertical tracking error. At $-12~\mathrm{m/s}$, it maintains good accuracy in the $z$-direction, though a noticeable 
bias in the negative $x$-direction appears as the horizontal wind becomes stronger. 
Notably, at this highest wind speed, the configuration combining the learned 
BEMT-based compensator with the rotor-speed allocation scheme achieves the lowest 
tracking error in both the $x$- and $z$-directions among all methods, indicating better generalization to unseen wind conditions than DAIML.

In the circular wall-contact scenario (Table~\ref{tab:tracking_comparison1}), the configuration without a compensator maintains acceptable tracking performance under a $-4~\mathrm{m/s}$ wind, with stable wall contact and moderate normal force. At $-8~\mathrm{m/s}$, however, the vehicle begins to drift in the negative $z$-direction and deviates from the wall’s symmetry plane ($y=0$), producing an 
elongated trajectory in the $y$-direction. The associated reduction and variation in normal contact force further reflect the controller’s reduced ability to regulate the wall-contact interaction. At $-12~\mathrm{m/s}$, the uncompensated controller is unable to maintain stable wall contact and to stay on the reference trajectory. The unusually large RMS values in this test condition are not physically meaningful, as the simulation is driven outside the valid operating range of the aerodynamic model. The DAIML configuration remains effective at $-4$ and $-8~\mathrm{m/s}$, exhibiting accurate tracking with consistent normal force regulation. At $-12~\mathrm{m/s}$, DAIML continues to follow the circular trajectory but shows noticeable deformation in the $y$-direction and greater variation in contact force, similar to the uncompensated configuration at $-8~\mathrm{m/s}$.
The proposed BEMT-based method achieves reliable tracking across all wind conditions. At $-4~\mathrm{m/s}$, it converges to the reference trajectory shortly after a brief transient with a slightly larger tracking error compared to DAIML. After the transient stage, DAIML and proposed configuration share a very similar tracking and contact force variation. At $-12~\mathrm{m/s}$, the combined BEMT-based compensator and rotor-speed allocation achieve the lowest RMS tracking error in both the $x$- and $z$-directions, while also maintaining the most stable wall-contact force. The lateral deformation in the $y$-direction is significantly reduced relative to DAIML, and the force trace displays fewer peaks, indicating improved contact stability. 
Overall, both DAIML and the proposed method maintain strong tracking performance at $-4$ and $-8~\mathrm{m/s}$, whereas the controller without compensation degrades substantially. Under the most challenging $-12~\mathrm{m/s}$ wind condition, the proposed method preserves continuous wall contact, maintains the most stable force profile, and achieves the highest tracking accuracy despite the compounded aerodynamic and contact-induced disturbances, demonstrating the best disturbance-rejection capability.

\section{Conclusion} \label{sec:conclu}

This paper presented a physics-infused disturbance-rejection framework for aerial
manipulators operating under strong wind disturbances and in close proximity to
vertical structures. The framework integrates three key components: a nominal
aerodynamic model constructed using Blade Element Momentum Theory (BEMT) with
parameters learned from free-flight data; a learning-based residual force model
trained to capture unmodeled aerodynamic effects; and an online adaptive
disturbance observer that compensates for residual aerodynamic mismatch as well as
contact-induced forces. In parallel, a BEMT-based rotor-speed allocation scheme
refines the thrust command at the actuation level by accounting for wind-dependent
rotor–thrust variation. Together, these components provide a structured and 
data-efficient approach to aerodynamic disturbance estimation and compensation.

\begin{table}[H]
\centering
\scriptsize
\begin{threeparttable}
\caption{Figure-eight tracking performance}
\label{tab:tracking_comparison}

\begin{tabular}{>{\centering\arraybackslash}p{1.2cm} c c c}
\toprule
\shortstack{\textbf{Wind}\\\textbf{[m/s]}} &
\shortstack{\textbf{No}\\\textbf{Compensator}} &
\textbf{DAIML~\cite{o2022neural}} &
\shortstack{\textbf{Proposed}\\\textbf{BEMT-Based Method}} \\
\midrule

$-4$ &
\raisebox{-.5\height}{\includegraphics[width=0.25\linewidth]{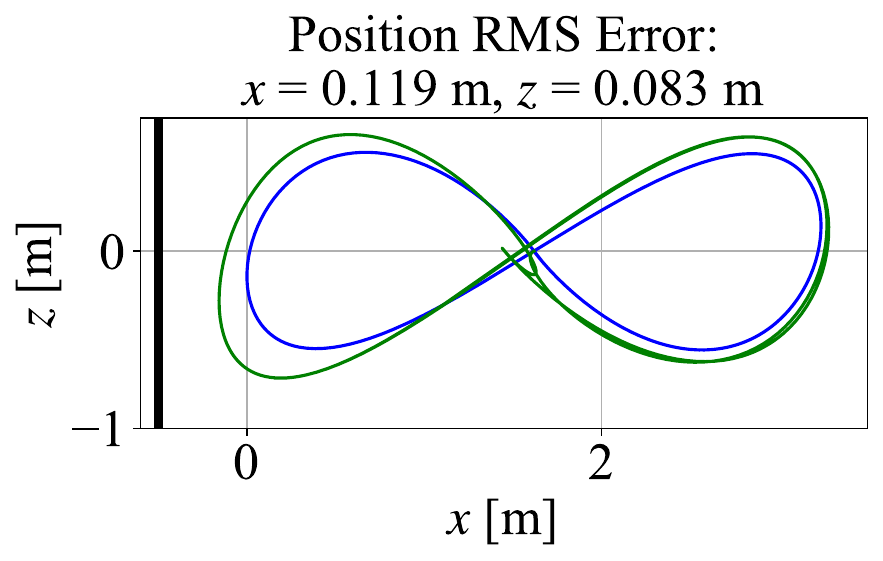}} &
\raisebox{-.5\height}{\includegraphics[width=0.25\linewidth]{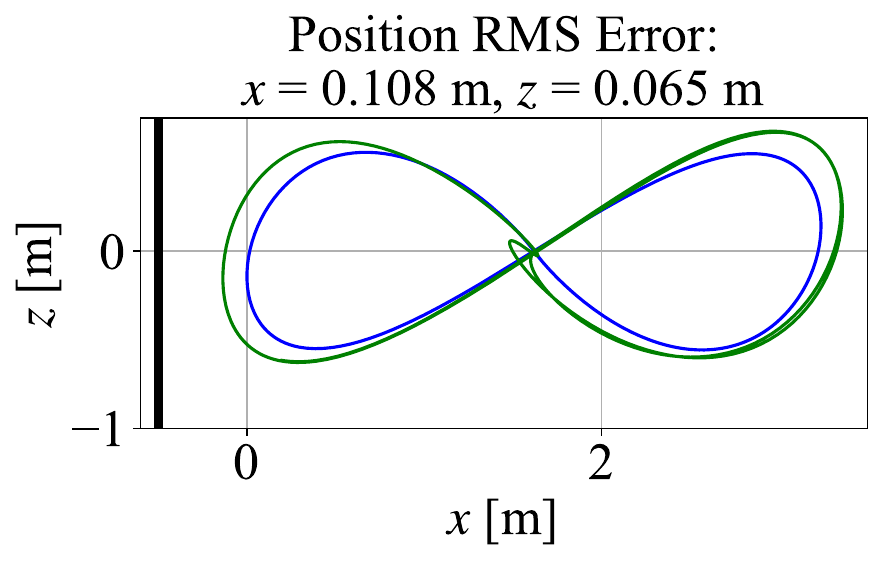}} &
\raisebox{-.5\height}{\includegraphics[width=0.25\linewidth]{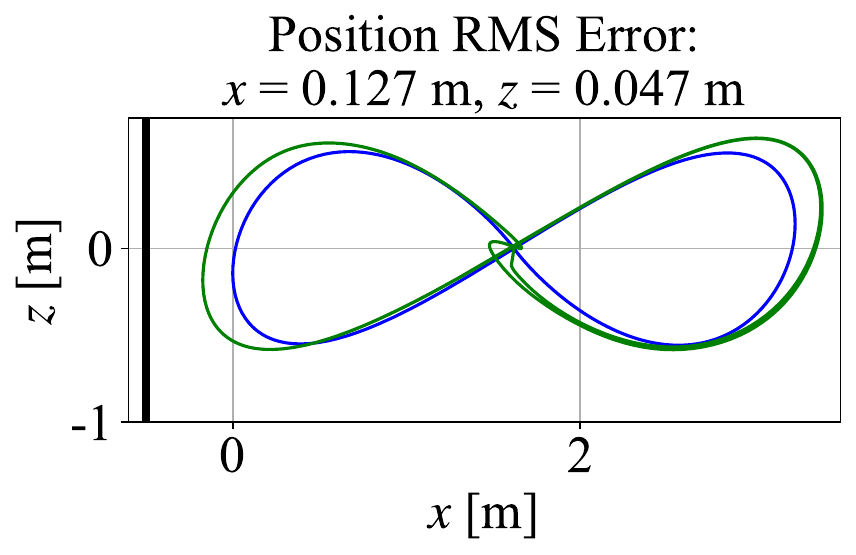}} \\

$-8$ &
\raisebox{-.5\height}{\includegraphics[width=0.25\linewidth]{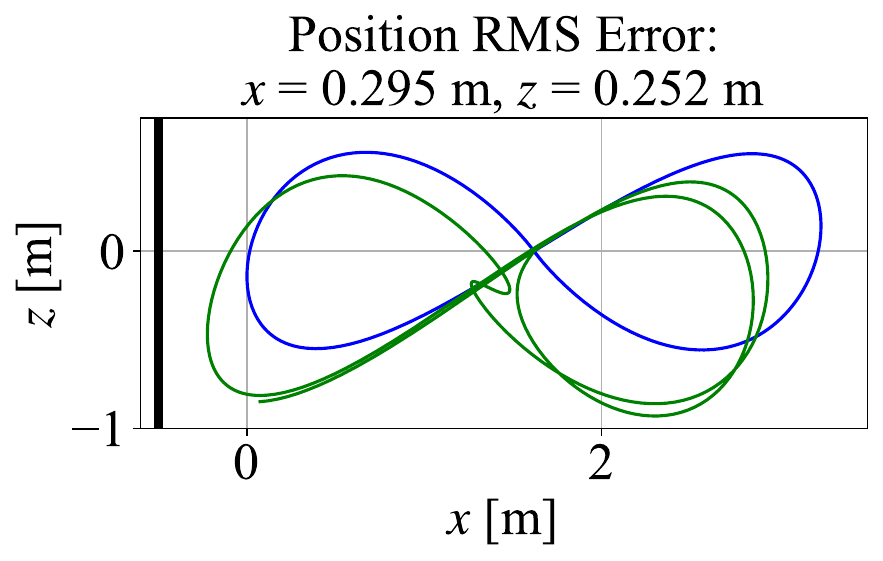}} &
\raisebox{-.5\height}{\includegraphics[width=0.25\linewidth]{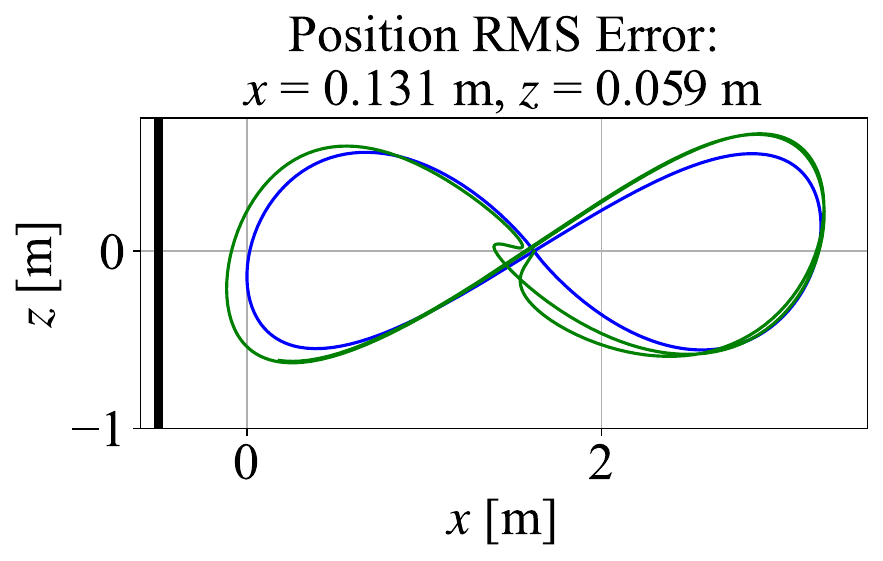}} &
\raisebox{-.5\height}{\includegraphics[width=0.25\linewidth]{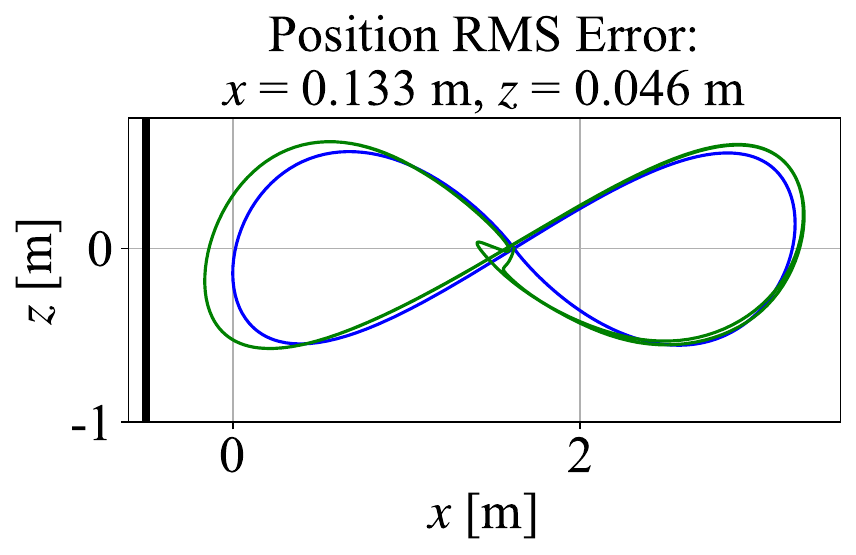}} \\

$-12$ &
\raisebox{-.5\height}{\includegraphics[width=0.25\linewidth]{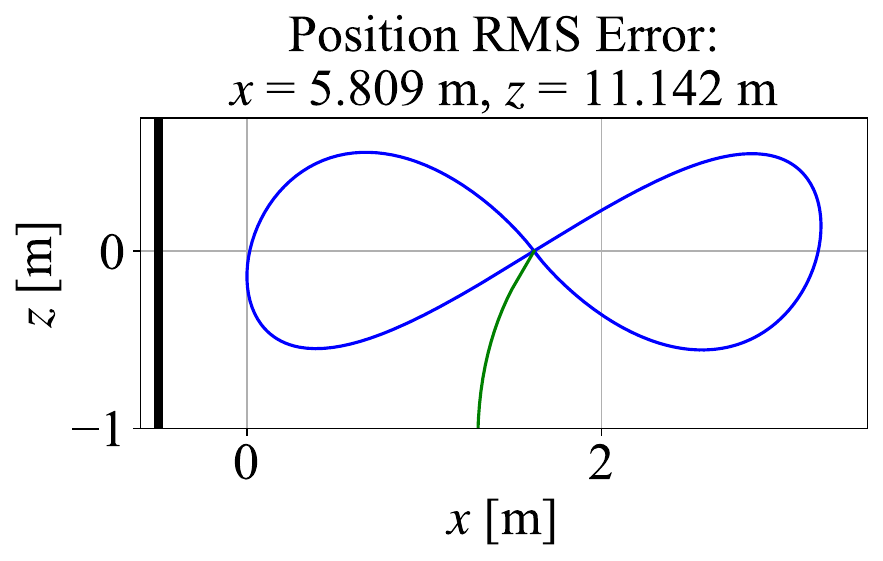}} &
\raisebox{-.5\height}{\includegraphics[width=0.25\linewidth]{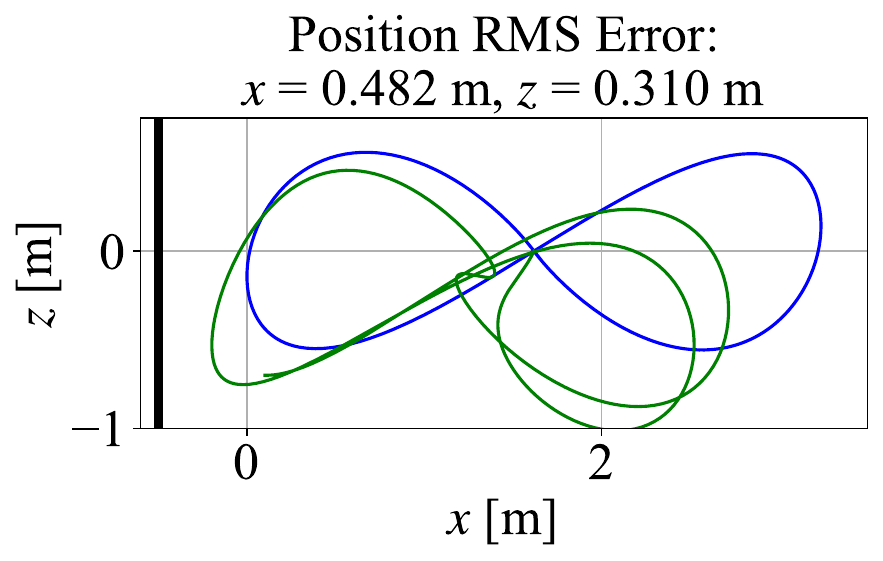}} &
\raisebox{-.5\height}{\includegraphics[width=0.25\linewidth]{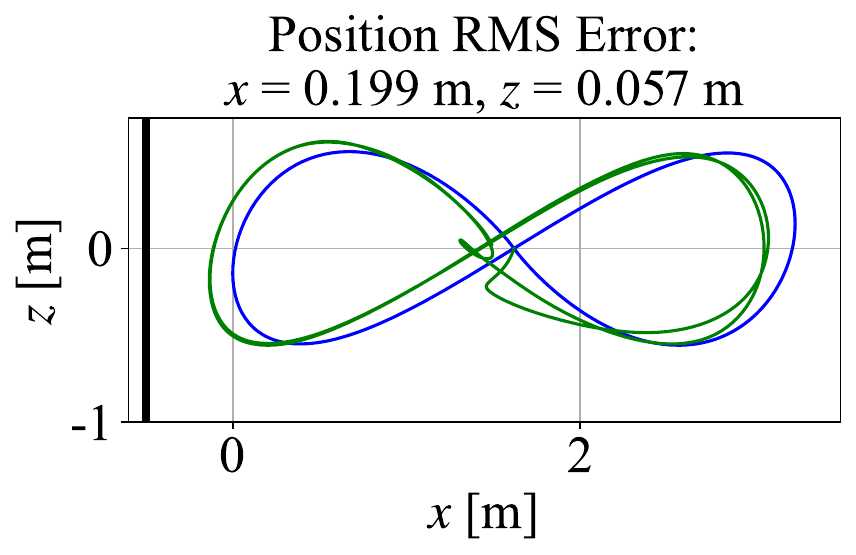}} \\

\bottomrule
\end{tabular}

\begin{tablenotes}
\footnotesize
\item Wind values shown in the first column are applied equally along the $x$- and $z$-directions.
\end{tablenotes}

\end{threeparttable}
\end{table}

\begin{table}[H]
\centering
\scriptsize
\setlength{\tabcolsep}{3pt}  
\begin{threeparttable}
\caption{Wall-contact tracking performance}
\label{tab:tracking_comparison1}

\begin{tabular*}{\textwidth}{@{}
  >{\centering\arraybackslash}p{0.8cm}   
  >{\centering\arraybackslash}p{0.28\textwidth}   
  >{\centering\arraybackslash}p{0.28\textwidth}   
  >{\centering\arraybackslash}p{0.28\textwidth}   
  >{\centering\arraybackslash}m{0.2cm}            
}
\toprule
\shortstack{\textbf{Wind}\\\textbf{[m/s]}} &
\shortstack{\textbf{No}\\\textbf{Compensator}} &
\textbf{DAIML~\cite{o2022neural}} &
\shortstack{\textbf{Proposed}\\\textbf{BEMT-Based Method}} &
\\
\midrule

$-4$ &
\raisebox{-.5\height}{\includegraphics[width=\linewidth]{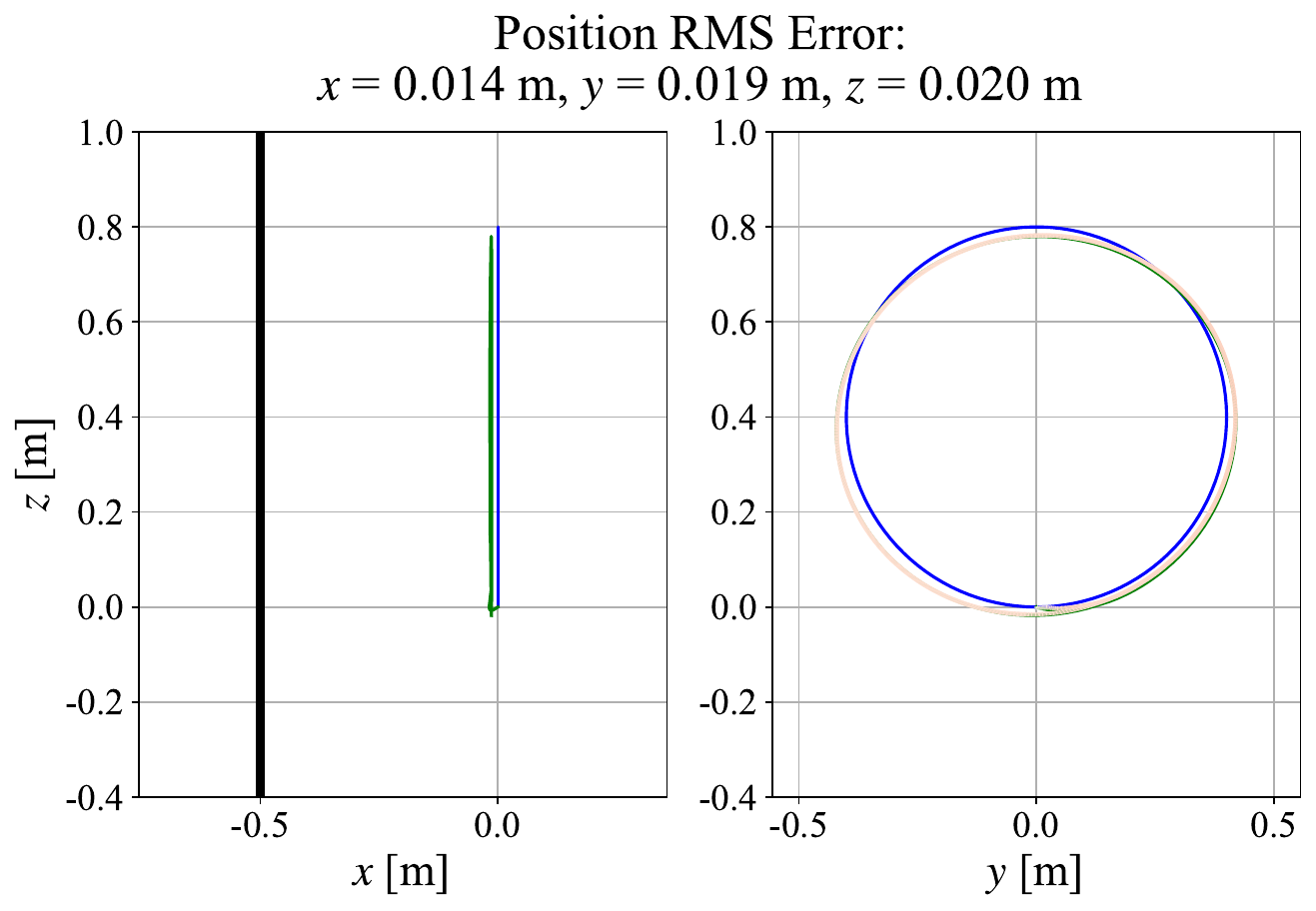}} &
\raisebox{-.5\height}{\includegraphics[width=\linewidth]{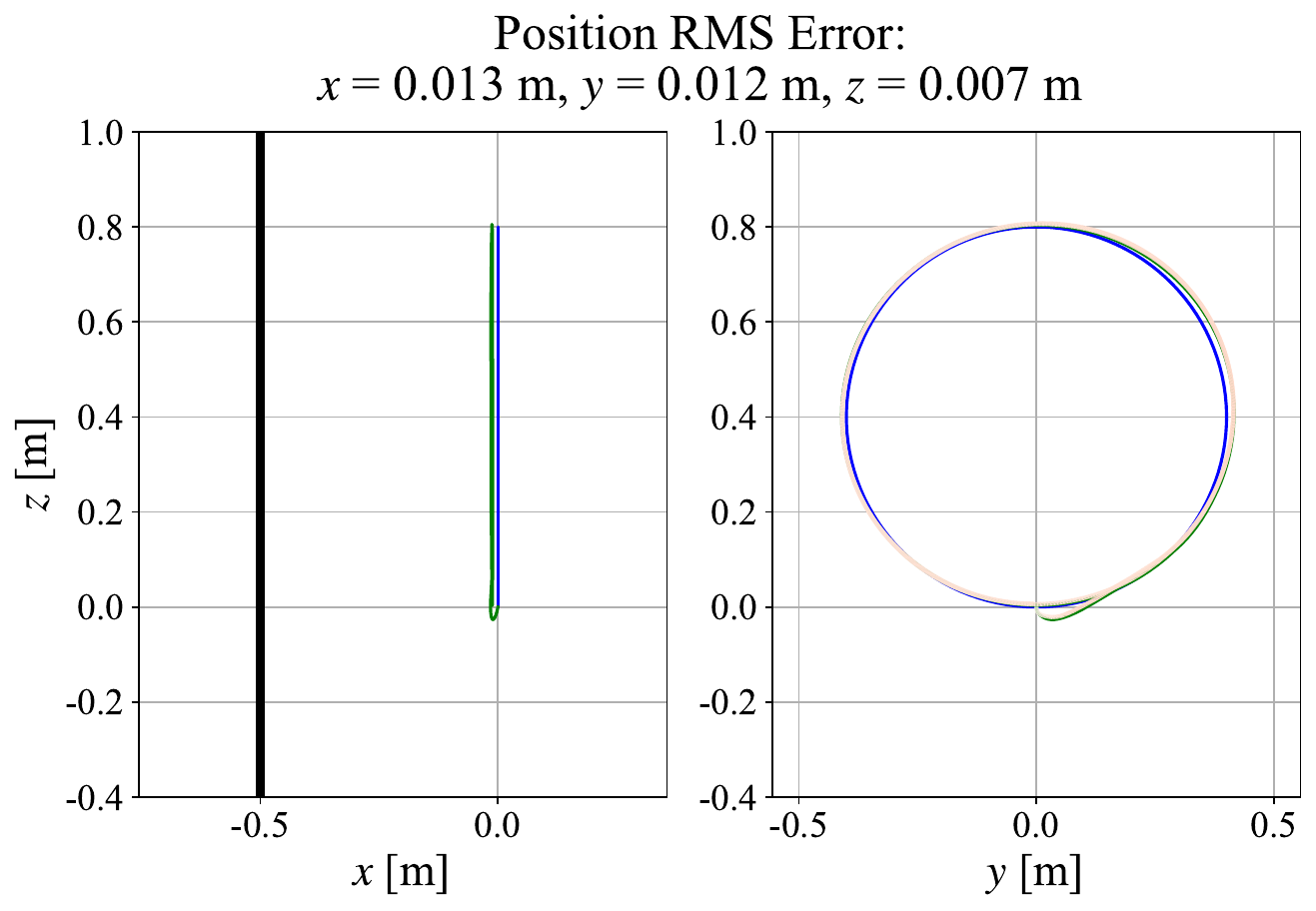}} &
\raisebox{-.5\height}{\includegraphics[width=\linewidth]{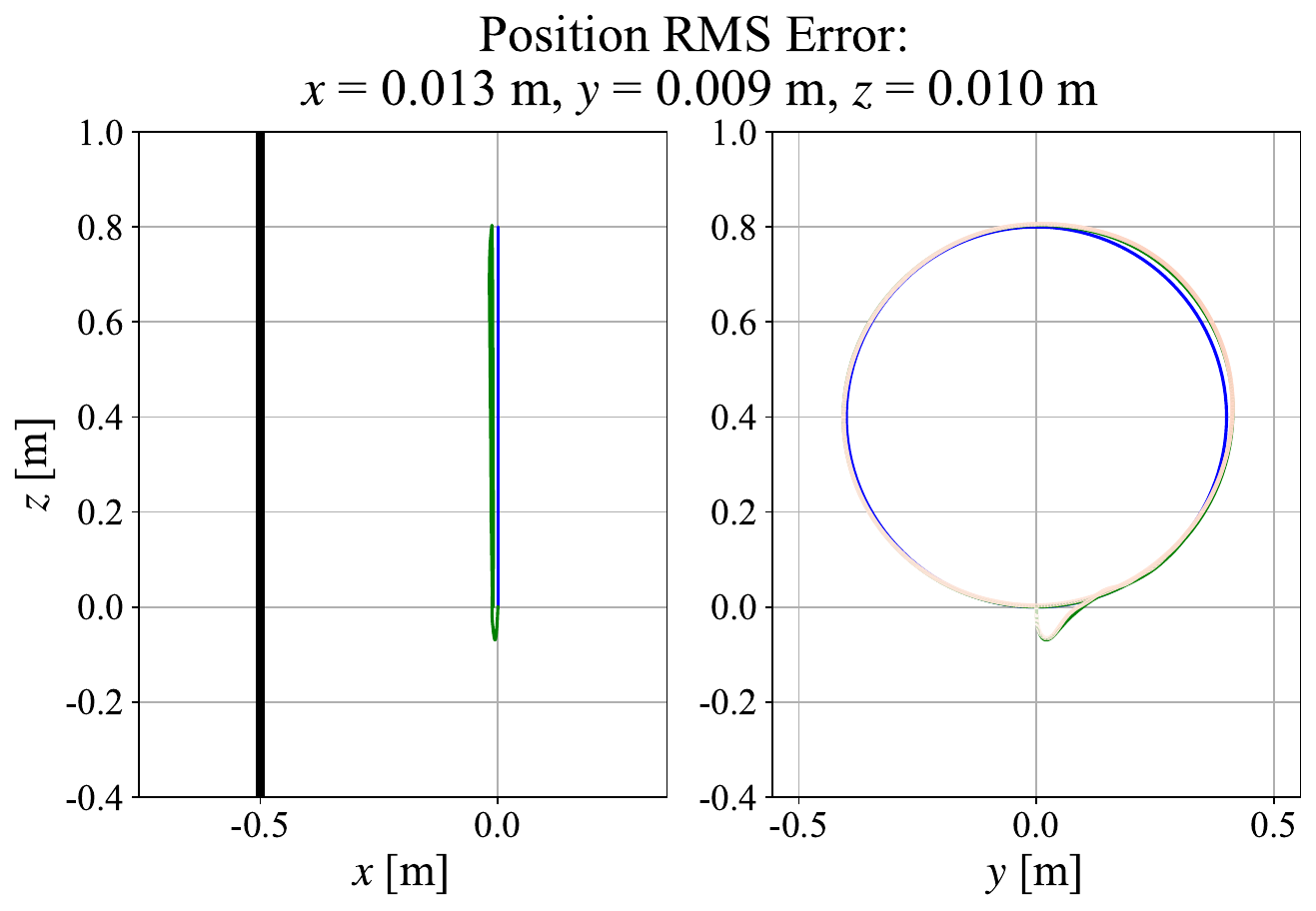}} &
\multirow{3}{*}{
  \raisebox{1.0\height}{
    \includegraphics[height=6.0cm]{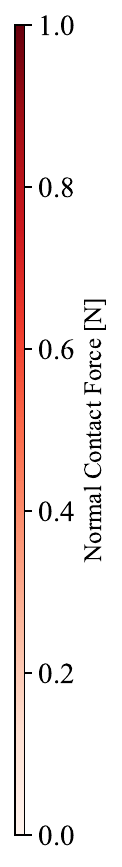}
  }
}
\\

$-8$ &
\raisebox{-.5\height}{\includegraphics[width=\linewidth]{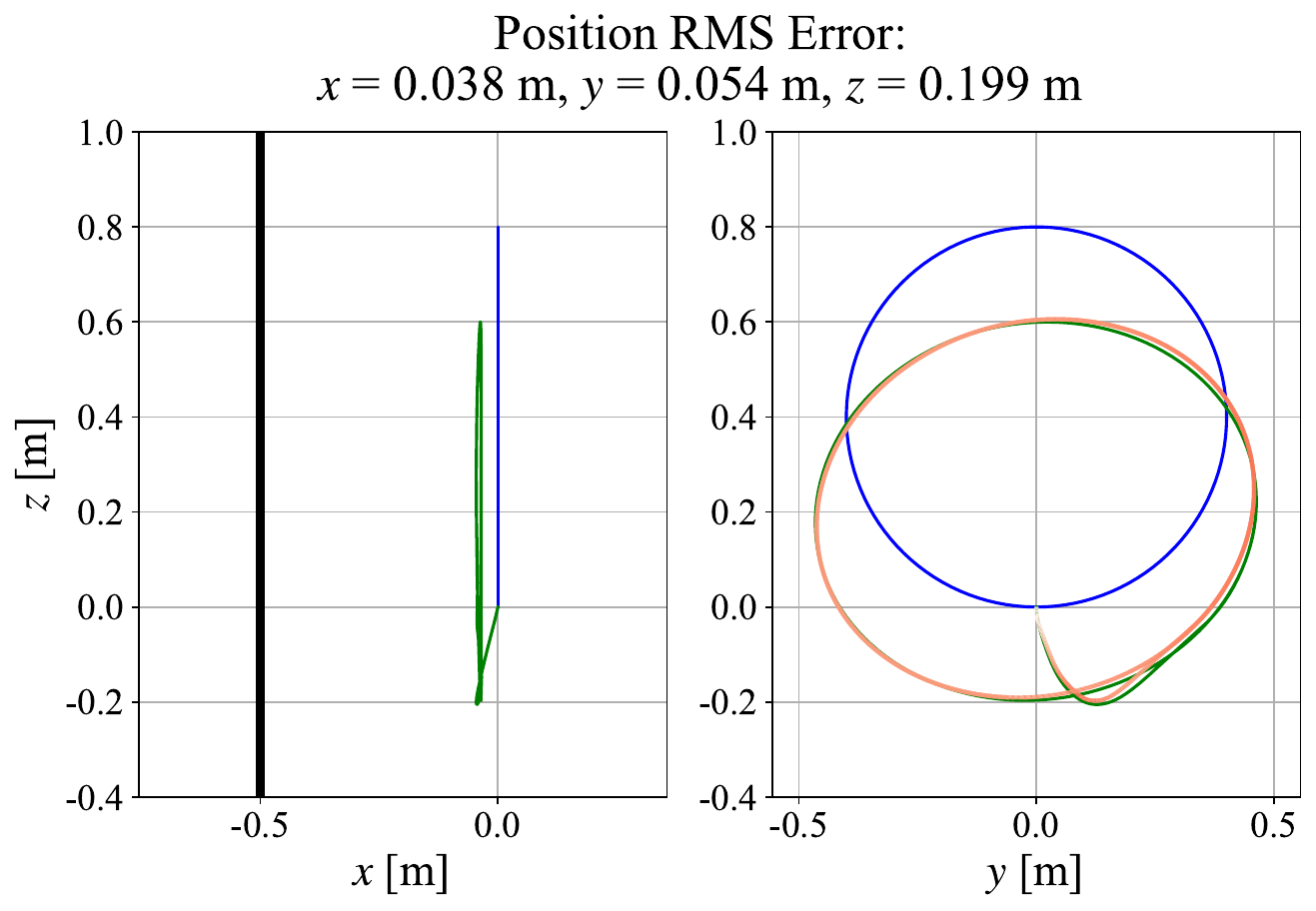}} &
\raisebox{-.5\height}{\includegraphics[width=\linewidth]{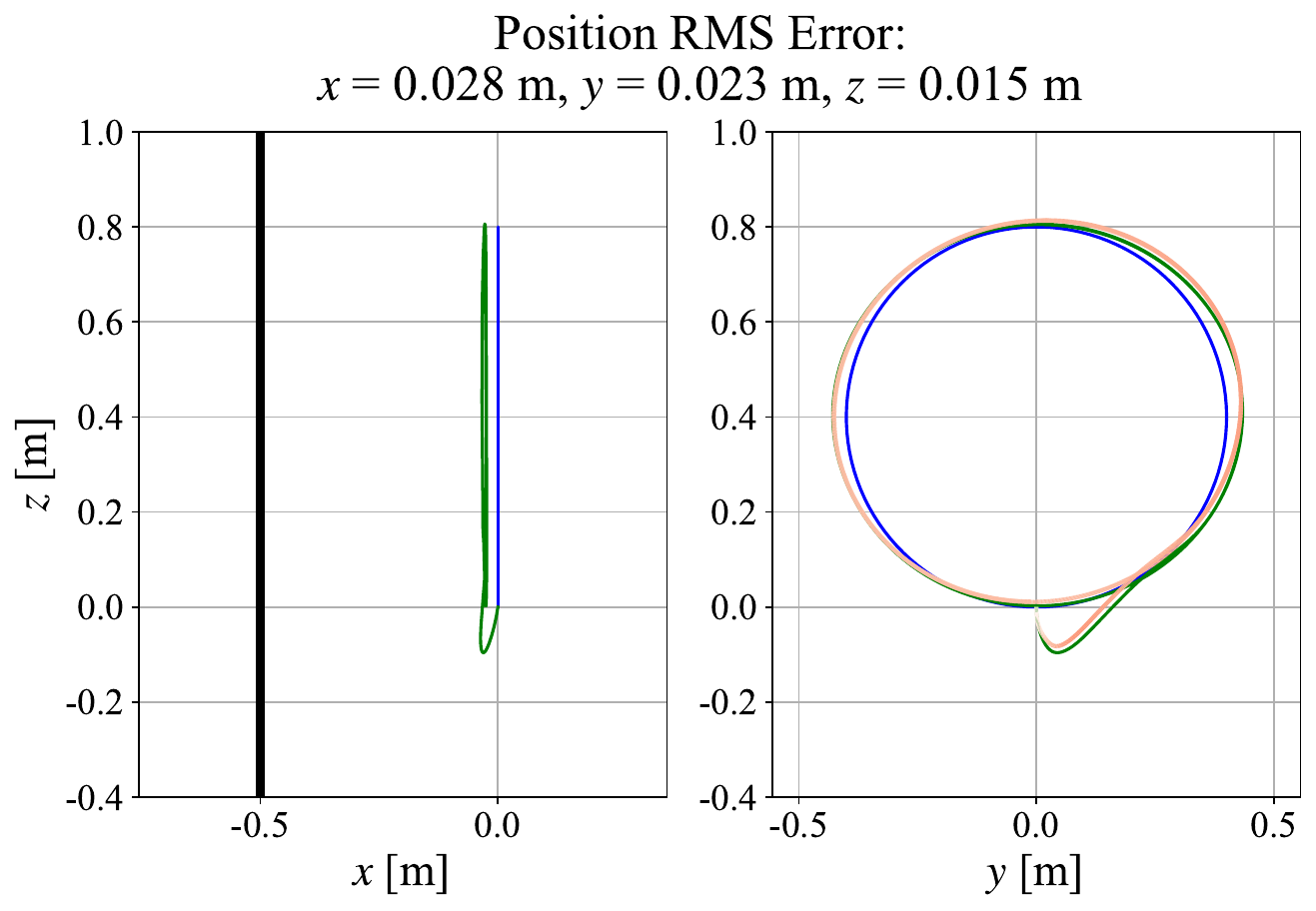}} &
\raisebox{-.5\height}{\includegraphics[width=\linewidth]{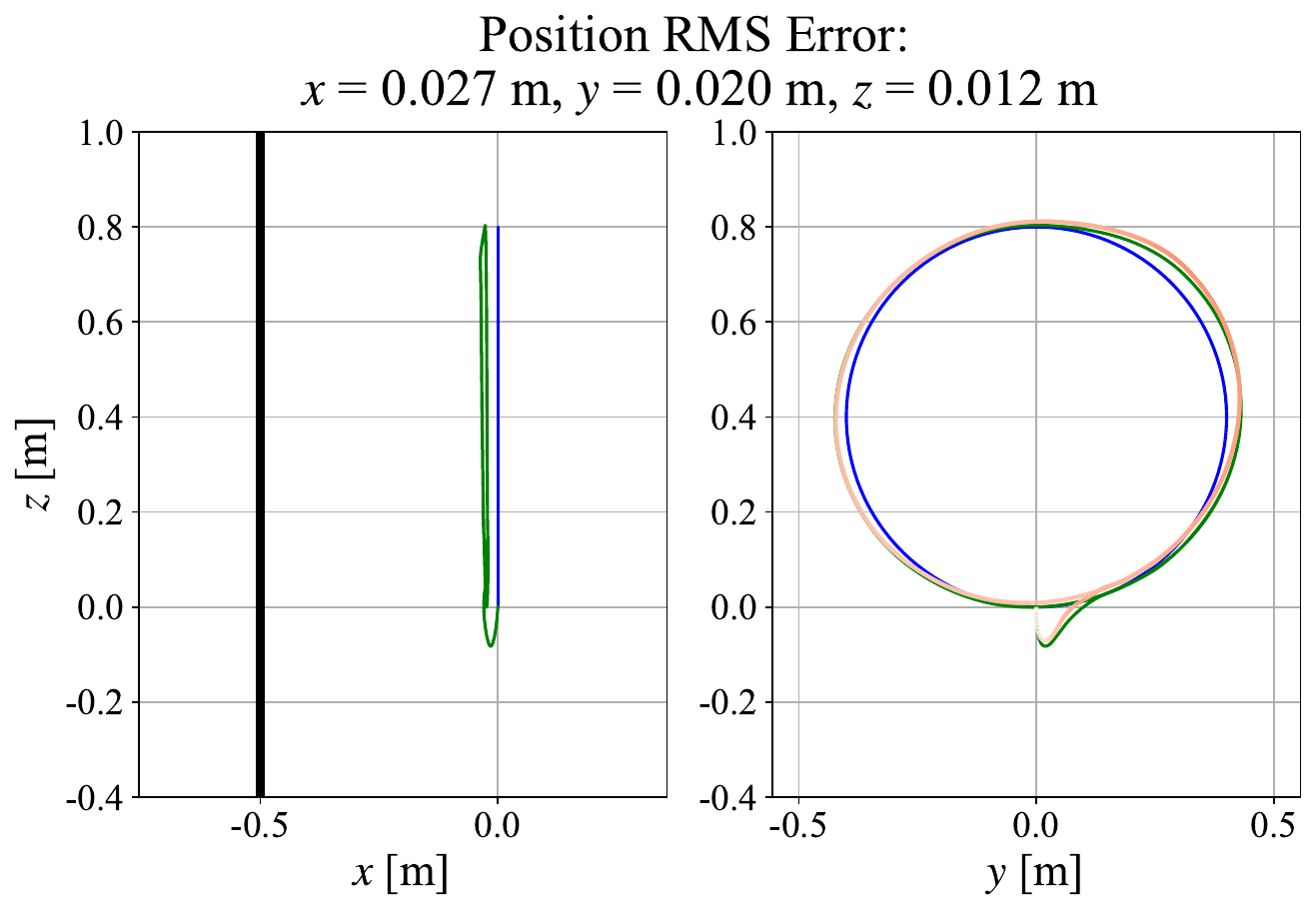}} &
\\

$-12$ &
\raisebox{-.5\height}{\includegraphics[width=\linewidth]{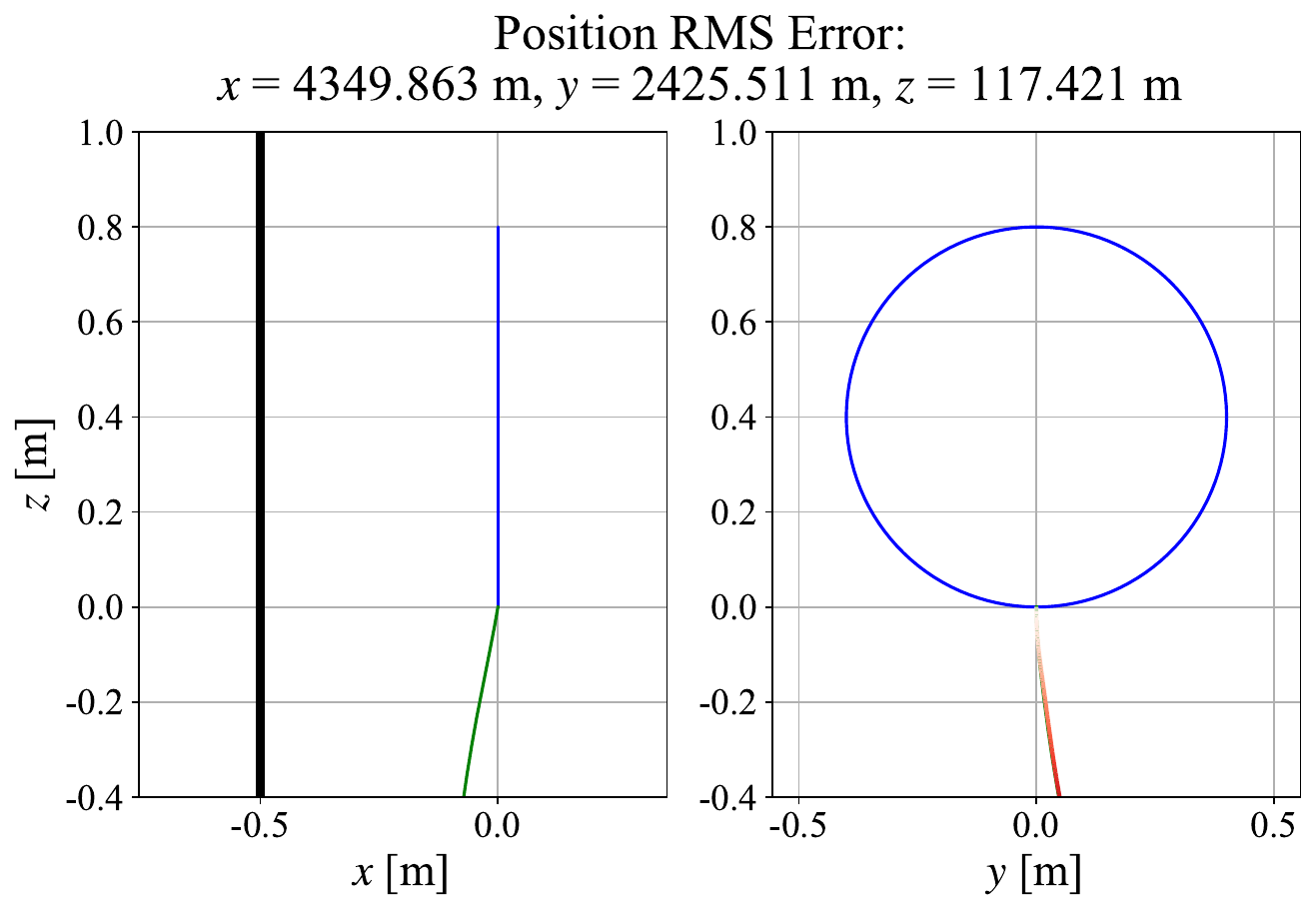}} &
\raisebox{-.5\height}{\includegraphics[width=\linewidth]{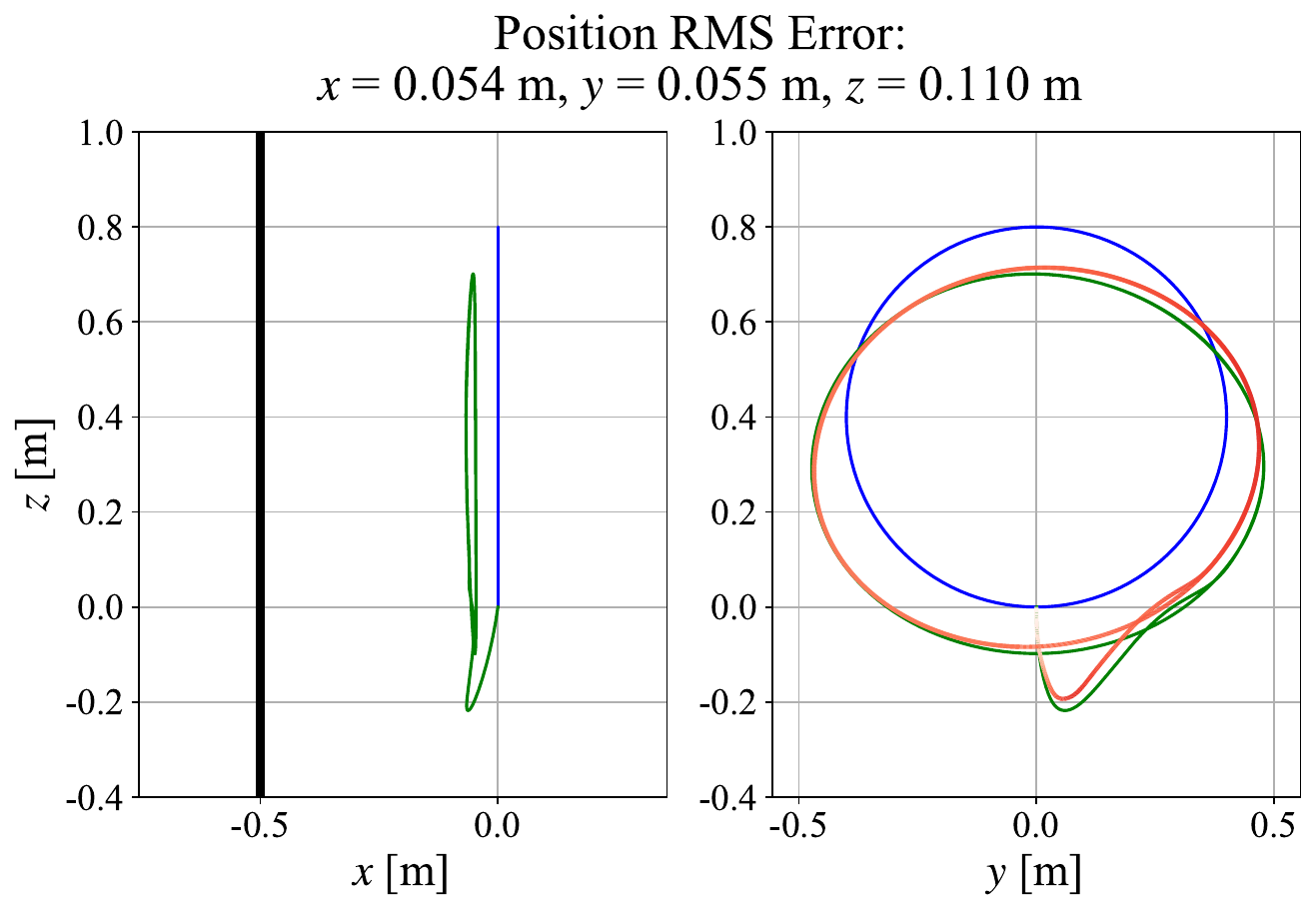}} &
\raisebox{-.5\height}{\includegraphics[width=\linewidth]{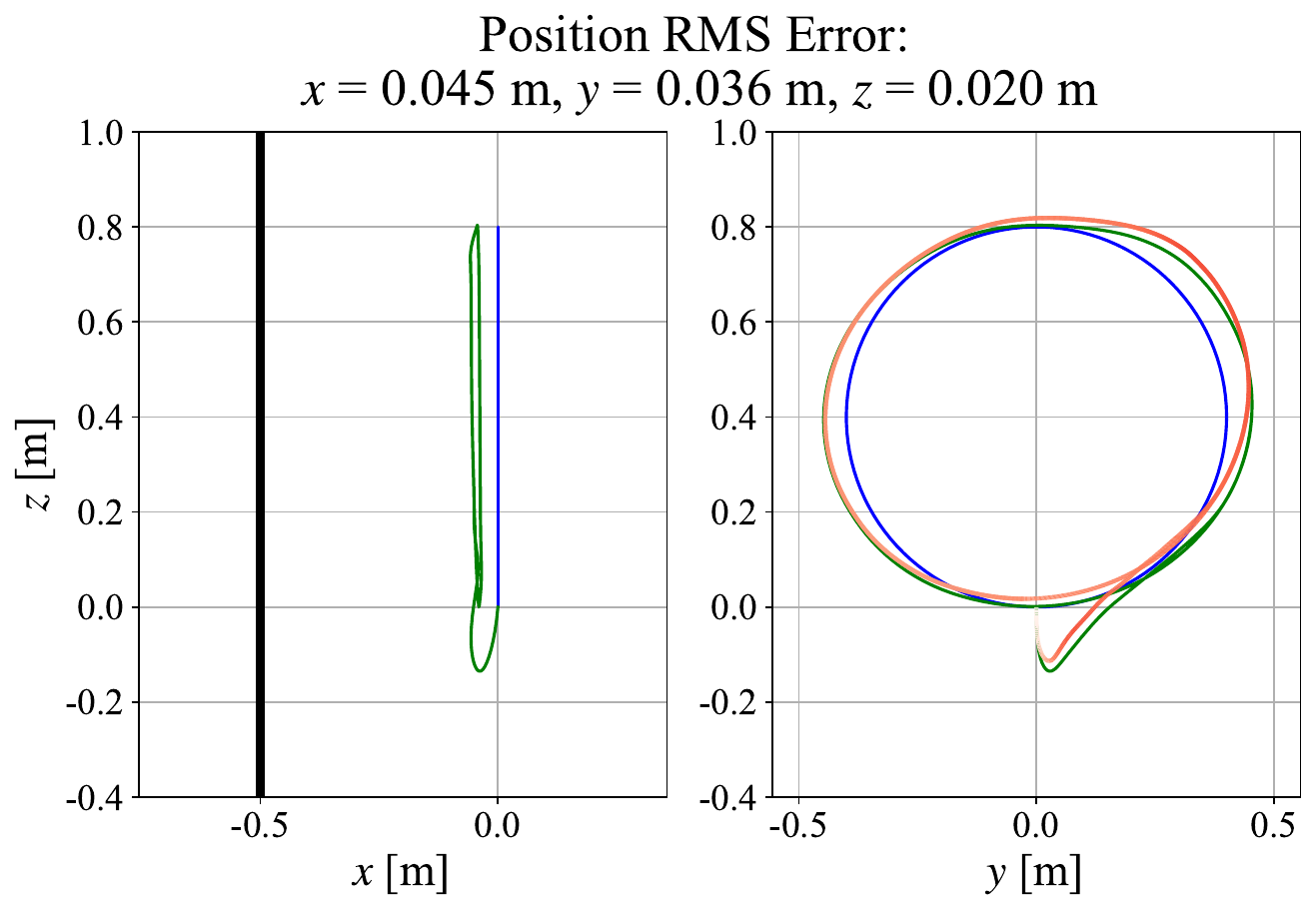}} &
\\

\bottomrule
\end{tabular*}

\begin{tablenotes}
\footnotesize
\item Wind values shown in the first column are applied equally along the $x$- and $z$-directions.
\end{tablenotes}

\end{threeparttable}
\end{table}

The proposed method was evaluated in a customized simulation environment that incorporates spatially varying wind fields, refined BEMT-based ground truth generation, and wall-contact interactions modeled through a deformable end-effector.
In the free-flight figure-eight task, all controllers maintained comparable performance under a $-4~\mathrm{m/s}$ wind. As wind intensity increased to $-8~\mathrm{m/s}$, the configuration without a compensator exhibited significant degradation, while the DAIML baseline and the proposed method both sustained stable tracking. At $-12~\mathrm{m/s}$, DAIML no longer generalized effectively and displayed growing tracking errors, whereas the proposed controller achieved the smallest RMS errors in both the horizontal and vertical directions, demonstrating superior robustness to unseen wind conditions.

A similar pattern emerged in the circular wall-contact task, where the contact force
acts as an additional unmodeled disturbance. Both DAIML and the proposed method
maintained stable wall contact and accurate tracking at $-4$ and $-8~\mathrm{m/s}$,
while the uncompensated controller degraded substantially. Under the most 
challenging $-12~\mathrm{m/s}$ condition, the proposed method achieved the highest
tracking accuracy and the most stable contact-force regulation, significantly
mitigating the combined aerodynamic and contact-induced disturbances.

These results highlight the benefits of combining high-fidelity aerodynamic
modeling with learned residual dynamics and adaptive estimation. The proposed
framework improves disturbance awareness at both the model and actuation levels,
leading to consistent performance across widely varying wind conditions and during
physical interaction with the environment. This approach provides a promising step
toward robust aerial manipulation in realistic, unstructured, and
aerodynamically complex environments.

\bibliography{references}.
\newpage
\section*{Appendix}
\label{sec:appendix}

\subsection{BEMT Force and Torque Computation Algorithm}
\label{app:bemt}

\begin{algorithm}[H]
\caption{Computation of $\mathbf{F}_{\mathrm{BEMT}}$ and $\mathbf{T}_{\mathrm{BEMT}}$}
\label{alg:bemt}
\begin{algorithmic}[1]

\Require Rotor speed, blade geometry, airfoil lift/drag curves, local wind velocity

\State \textbf{Initialize} an induced inflow estimate $v_i(j)$ for each radial blade element $j$.

\Statex
\For{each radial blade element $j$} \label{line:for-j}

    \Repeat  \label{line:repeat-j}

        \State $T_{\mathrm{BET}} \gets 0$

        \For{each azimuthal (rotational) segment $k$}

            \State Compute local 3D airflow at section $(j,k)$ from rotor speed, wind, and $v_i(j)$.
            \State Project this airflow onto the blade-sectional plane.
            \State Compute angle of attack and the local relative flow magnitude.
            \State Evaluate lift and drag coefficients from the airfoil model.
            \State Form elemental lift and drag vectors in the section frame and transform them
                   to the rotor-disk frame.
            \State Accumulate the elemental thrust contribution into $T_{\mathrm{BET}}(v_i(j))$.

        \EndFor

        \State Compute the momentum-theory thrust $T_{\mathrm{MT}}(v_i(j))$ for this radial annulus.
        \State Update $v_i(j)$ with a one-dimensional root-finding step to reduce
               $\lvert T_{\mathrm{BET}}(v_i(j)) - T_{\mathrm{MT}}(v_i(j)) \rvert$.

    \Until{convergence of $v_i(j)$ for element $j$} \label{line:until-j}

    \State Recompute elemental lift and drag over azimuth using the converged $v_i(j)$.
    \State Using the same azimuthal summation, compute the elemental axial torque about the rotor hub.
    \State Accumulate this radial element's contribution to the net rotor force and axial rotor torque.

\EndFor \label{line:endfor-j}

\Statex
\State Integrate the accumulated radial contributions over the blade span to obtain the rotor aerodynamic force $\mathbf{F}_{\mathrm{BEMT}}$ and the net axial torque.
\State Form the rotor aerodynamic torque vector $\mathbf{T}_{\mathrm{BEMT}}$ by placing the axial torque in the shaft direction and setting the lateral components to zero.

\Statex
\State \Return $\mathbf{F}_{\mathrm{BEMT}}$ and $\mathbf{T}_{\mathrm{BEMT}}$

\end{algorithmic}
\end{algorithm}

\subsection{Definition of the Composite Tracking Error}
\label{app:s}
The tracking errors consist of both translational and rotational components. Let $\mathbf{x}\in\mathbb{R}^3$
denote the vehicle position and $\mathbf{x}_r\in\mathbb{R}^3$ the reference position. The corresponding translational velocity is
$\mathbf{v}\in\mathbb{R}^3$, with reference velocity $\mathbf{v}_r$.
The translational tracking errors are therefore
\begin{align}
    \mathbf{e}_x &= \mathbf{x} - \mathbf{x}_r, \\
    \mathbf{e}_v &= \mathbf{v} - \mathbf{v}_r.
\end{align}

The attitude of the vehicle is represented by the rotation matrix $\mathbf{R} \in \mathrm{SO}(3)$. The reference attitude is denoted by $\mathbf{R}_r \in \mathrm{SO}(3)$ and has the same frame interpretation.
Using these matrices, the attitude tracking error is defined through the vee map $(\cdot)^\vee$, which extracts the vector associated with a skew-symmetric matrix:
\begin{equation}
    \mathbf{e}_R
    = \tfrac{1}{2}\!\left(\mathbf{R}_r^\top \mathbf{R}
    - \mathbf{R}^\top \mathbf{R}_r\right)^\vee.
\end{equation}
Let $\boldsymbol{\omega}$ denote the body-frame angular velocity and
$\boldsymbol{\omega}_r$ its reference. The corresponding angular-velocity
tracking error is
\begin{equation}
    \mathbf{e}_\omega
    = \boldsymbol{\omega}
    - \mathbf{R}^\top \mathbf{R}_r\,\boldsymbol{\omega}_r.
\end{equation}
We define the stacked error vectors
\[
    \tilde{\mathbf{x}} = 
    \begin{bmatrix}
        \mathbf{e}_x \\
        \mathbf{e}_R
    \end{bmatrix},
    \qquad
    \tilde{\mathbf{v}} =
    \begin{bmatrix}
        \mathbf{e}_v \\
        \mathbf{e}_\omega
    \end{bmatrix},
\]
which collect the full translational and rotational deviations into a single
6-dimensional representation. To construct a first-order–like error variable
suitable for the $6$-D rigid-body dynamics, we introduce the composite tracking
error
\begin{equation}
    s = \tilde{\mathbf{v}} + \Lambda\,\tilde{\mathbf{x}},
    \label{eq:s_def}
\end{equation}
where $\Lambda \in \mathbb{R}^{6\times 6}$ is a positive-definite gain matrix.

\end{document}